%% file: main.tex
\documentclass[letterpaper,10pt]{IEEEtran}
\usepackage[margin=1in]{geometry}
\usepackage{amsmath} 
\usepackage{subcaption}
\usepackage{custom_commands} 
\usepackage[normalem]{ulem}

\usepackage[compress]{cite}
\usepackage{url}
\usepackage{graphicx,xcolor}
\usepackage{verbatim}
\makeatletter
\newcommand{\verbatimfont}[1]{\def\verbatim@font{#1}}%
\makeatother
\verbatimfont{\ttfamily\small}

\newcommand{\bi}{\begin{itemize}}\newcommand{\ei}{\end{itemize}}
\newcommand{\be}{\begin{equation}}\newcommand{\ee}{\end{equation}}
\newcommand{\bee}{\begin{enumerate}}\newcommand{\eee}{\end{enumerate}}
\newcommand{\bea}{\begin{eqnarray}}\newcommand{\eea}{\end{eqnarray}}
\newcommand{\beas}{\begin{eqnarray*}}\newcommand{\eeas}{\end{eqnarray*}}
\newcommand{\bc}{\begin{center}}\newcommand{\ec}{\end{center}}

\newcommand{\strike}[1]{}
\newcommand{\add}[1]{#1}
\newcommand{\strikeb}[1]{}
\newcommand{\addb}[1]{#1}

\newcommand{\useSD}[1]{res_sd/#1}

\title{Feedback Control of an Exoskeleton for Paraplegics\\
\Large Toward Robustly Stable Hands-free Dynamic Walking}
\author{Omar Harib,
Ayonga Hereid,
Ayush Agrawal,
Thomas Gurriet, \\
Sylvain Finet,
Guilhem Boeris,
Alexis Duburcq,
M. Eva Mungai, \\
Matthieu Masselin,
Aaron D. Ames,
Koushil Sreenath,
Jessy Grizzle\\
	POC: O.\ Harib(oharib@umich.edu) }

\newif\ifPDF \ifx\pdfoutput\undefined\PDFfalse \else\ifnum\pdfoutput > 0\PDFtrue \else\PDFfalse \fi \fi
\ifPDF 
\usepackage[pdftex, plainpages = false, colorlinks=true, linkcolor=black, citecolor = green!50!blue, urlcolor = blue, filecolor=black, pagebackref=false, hypertexnames=false,  pdfpagelabels ]{hyperref}

\newcommand{\real}{{\mathbb{R}}}

\begin{document}

\maketitle

\input{section_introduction}
\input{section_model}

\input{section_ghzd_controller_overview}
\input{section_phzd}

\input{section_optim_plus_machinelearning.tex}

\input{section_numerical_validation}
\input{section_phzd_experiments}
\input{section_conclusion}

\section*{Acknowledgments}
\add{The US investigators are covered under IRB protocol number 16-0693. The French investigators are covered under CPP Ile de France XI,  number 16041.}

The authors thank the entire Wandercraft team who designed ATALANTE as well as implemented and tested the control algorithms used in the experiments. Additionally, the authors are grateful to Laurent Praly and Nicolas Petit, researchers at CAS (Mines ParisTech, PSL Research University) for their scientific support since the founding of Wandercraft. The authors thanks Xingye (Dennis) Da for his consultation on GHZD.\\

The work of A. Hereid was supported by NSF under Grant CPS-1239037. The work of O. Harib and J. W. Grizzle was supported in part by NSF Grant NRI-1525006  and in part by a gift from Ford Motor Company. M.E. Mungai is supported by a University of Michigan Fellowship.  The work of A. Agrawal and K. Sreenath is supported by NSF grant IIS-1526515. The work of Thomas Gurriet and Aaron Ames was supported by NSF NRI award 1526519.



\appendix


\input{sidebar_pinned_vs_floating.tex}


\input{sidebar_direct_collocation.tex}


\input{sidebar_cybalathon.tex}

\clearpage
\bibliography{exo,sidebar} 
\bibliographystyle{IEEEtran}
\end{document}

%% file: section_introduction.tex
\begin{figure}
	\centering
    \includegraphics[width=0.8\columnwidth]{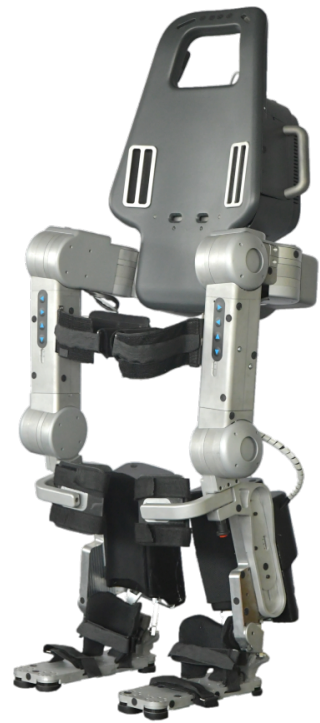} 
    \caption{ATALANTE: An exoskeleton designed by Wandercraft for people with paraplegia.}  
	\label{fig:atalante_v4_nobackground_blurred}
\end{figure}

``I will never forget the emotion of my first steps [...],'' Fran\c{c}oise, first user during initial trials of the exoskeleton ATALANTE \cite{WandercraftOnline} discussed in this paper. ``I am tall again!'', Sandy, the fourth user after standing up in the exoskeleton. In these early tests, complete paraplegic patients have dynamically walked up to 10 m without crutches or other assistance using a feedback control method originally invented for bipedal robots. This paper describes the hardware, shown in \figref{fig:atalante_v4_nobackground_blurred}, that has been designed to achieve hands-free dynamic walking, the control laws that have been deployed and those being developed to provide enhanced mobility and robustness, as well as the early test results alluded to above. \add{In this paper, dynamic walking refers to a motion that is orbitally stable as opposed to statically stable.}

\strike{Across the world, 70 million walking-impaired people have one life objective: to walk again.}
At present approximately 4.7 million people in the United States would benefit from an active lower-limb exoskeleton due to the effects of stroke, polio, multiple sclerosis, spinal cord injury, and cerebral palsy \cite{DoHe2008}. 
Moreover, by 2050, an estimated 1.5 million people in the United States will be living with a major lower-limb amputation \cite{ZiMaEpTrBr2008}. Such individuals expend up to twice the metabolic effort to walk at half the speed of able-bodied persons, experience higher-risk of falls, and have secondary pathological conditions such as osteoarthritis, back pain, and depression \cite{WaPeAnHi1976,PeDoFoRu1993,MiDeSpKo2001}.
Lower-limb exoskeletons serve as assistive devices by providing support and balance to wheelchair users and enabling them to perform normal ambulatory functions such as standing, walking and climbing stairs. Lower-limb exoskeletons have also been utilized for gait training and rehabilitation purposes.

More importantly, standing and walking with these assistive devices provides exceptional health benefits.  For instance, for paraplegics the benefits include improvement of blood circulation, of respiratory, urinary, and intestinal functions, as well as positive psychological effects \cite{krebs1998robot}, fundamentally improving their quality of life. For spinal cord injury (SCI) patients, the benefits include improved bone density, cardiorespiratory function, gastrointestinal function, sitting balance, and decreased pain and spasticity \cite{GeKa2017}.  \strike{It is thus our moral job to enable these people to stand and walk.}

The objective of the work described herein is to translate formal control design methodologies from bipedal robots to exoskeleton systems so as to achieve dynamic hands-free walking. This is a formidable problem as control of bio-mechatronic exoskeleton devices not only share many of the challenges of bipedal robot locomotion but also challenges introduced by the integration of an active human user. These challenges include, nonlinear, high degree-of-freedom hybrid dynamics,  workspace limitations, actuator constraints, unilateral ground contact forces, being robust to variations in the user's dynamical parameters such as mass and inertia, being able to handle interaction forces between the user and the device, and enforcing safety-critical constraints for the operation of the exoskeleton.

Over the years, several research groups and companies have begun responding to the need and benefits of exoskeletons.  Exoskeletons can be designed for human performance augmentation and as orthotic devices.
Exoskeletons for Human Performance Augmentation are designed to enhance the strength and physical capabilities of able-bodied users, to provide fatigue relief and protection to factory and construction workers, soldiers, and disaster relief workers, and to assist them in carrying heavy loads for prolonged periods of time. 
In contrast, orthotic devices are designed to assist and restore autonomy to individuals with physical impairments causing difficulty in walking. Orthotic devices are also designed for rehabilitation purposes - to provide gait training and therapy. A comprehensive review of the state-of-the art lower-limb exoskeletons can be found in \cite{Dollar2008,Young2017,Huo2016}. 

Early exoskeletons, developed by General Electric \cite{mosher1967handyman}, the University of Wisconsin \cite{Grundmann1977}, and the Mihailo Pupin Institute \cite{vukobratovic1974development,hristic1981new}, focused on human augmentation or  assistance, supporting multi-task capabilities such as walking, standing up from a seated position, sitting down, stepping over obstacles, and climbing stairs, all through pre-programmed motion patterns that were executed at the user's command.  These systems, however, were not very robust.
Turning to more recent devices, Ekso Bionics' robotic exoskeleton EksoGT \cite{EksoOnline} is primarily designed for use in clinical settings for rehabilitation and gait training for stroke and SCI patients. ReWalk \cite{ReWalkOnline,Zeilig2012} is another robotic lower-limb exoskeleton to enable patients with SCI to stand-up, walk, turn and climb stairs. The Hybrid Assistive Limb (HAL) \cite{Suzuki2007} developed at the University of Tsukuba, Japan and Cyberdyne \cite{cyberdyneOnline} provides locomotion assistance to physically challenged persons. These are high-degree-of-freedom exoskeletons that have multiple actuators at the hip, knee and ankle. A primary limitation, however, of the cited exoskeletons is that they require external support mechanisms such as crutches or canes for the user to maintain balance while walking with the device. While REX bionics' lower-limb exoskeleton provides hands-free functionality, it only allows slow static gaits with velocities on the order of 0.05 m/s.

While modern day \add{hardware for exoskeletons and prosthetics is} becoming lighter, stronger, and power-dense, the current approaches to control of powered \add{leg devices} are rudimentary and driven by finite-state machines with several phases such as swing, stance, heel-strike and toe-off \cite{JiVe2012}\add{, \cite[Fig.~9]{sup2008design}}, 
each with numerous tunable parameters that are specific to each user \cite{SiFeFiLiHa2013,simon2014configuring}, offering no formal guarantees of either stability or safety \cite{GrLeHaSe2014}, and typically require the use of additional aids such as crutches to be safely used \cite{Stickland2012,angold2011semi}.
\add{}
A general review of various control strategies for \add{lower-limb assistive robotics} is presented in \cite{JiVe2012,AnAl2012,Tucker2015}. In particular, low-level control strategies are either position-based \cite{SuMiKaHaSa2007,AgCoPeGo2007,GoSiSi2011,NeGuRi2009,FrSoPrBeRoCh2011} or torque / force-based \cite{JeCoKeFrMo2003,TsHaSa2011,UnPiWeBoMa2011, WoReBo2010,RoBeRuMaMoPo2007}, while a higher-level impedance or admittance controller is used to regulate human-device interaction forces.
This is in stark contrast to the surge in control technology for highly dynamic bipedal locomotion \cite{SrPaPoGr2013,FeWhXiAt2015,KuPeTe2014,DaVaTe2014,IMHC_VRC2013,MIT_VRC}, where tools are being developed that allow rapid design of gaits and model-based feedback controllers, that respect physical constraints of the system such as torque limits and joint speeds, while providing formal guarantees on stability, safety and robustness to uncertainties in the model and in the environment  \cite{Westervelt2007Feedback,griffin2015walking,Access2015_CLFQP,RSS2015_RobustCLF,Hereid2018Dynamic}.
If the control and design methodologies underlying advanced locomotion strategies for bipedal robots can be successfully translated to powered prostheses  and exoskeletons in a holistic and formal manner, the end result promises to be a new generation of wearable robotic devices that deliver the next level of stable, safe, and efficient mobility.

A new paradigm of control design is thus necessary to achieve dynamic hands-free exoskeleton walking, one that transcends current approaches involving state machines and \add{extensive gain tuning \cite{Quintero2018Continuous,Zhao2016Multicontact,Agrawal2017First,Gurriet2018Towards}}.
The heart of our approach involves virtual constraints and hybrid invariant manifolds. Virtual constraints are functional relations achieved on the generalized coordinates of the exoskeleton via feedback control; they provide a systematic means for coordinating limb motion and providing corrective actions to attenuate disturbances, without resorting to low-dimensional pendulum models. Indeed, the virtual constraints are designed herein on the basis of the 18 degree-of-freedom floating-base model of the exoskeleton and offline trajectory optimization. This approach is validated both numerically and in pre-clinical experimental testing, the latter enabling paraplegics to walk hands-free, with all control actions required for stable dynamic walking provided by an onboard controller.
We also present a recent generalization of virtual constraints that is based on a unique combination a fast offline trajectory optimization and machine learning, in tandem with online robust trajectory tracking. These newer techniques harness the power of modern optimization tools and are blazing the way for improved controller designs that deliver multiple walking speeds, turning, and enhanced robustness for exoskeleton locomotion.

In the remainder of the paper, we introduce the exoskeleton mechanism under study, construct a dynamic model for control design, and develop control objectives for achieving hands-free dynamic walking.  Following this, state-of-the-art techniques in bipedal control are summarized and how to translate a method based on virtual constraints to exoskeletons is highlighted. A generalization of virtual constraints is presented that combines offline trajectory optimization and machine learning to design stabilized gaits that can be robustly tracked online. \addb{Preliminary robustness and stability analysis of both control design approaches are numerically illustrated in simulation,} while pre-clinical tests are currently available only for the first method. In these \add{early} tests, \add{aimed at evaluating the viability of the hardware and the approaches to control design discussed in the current paper, fully} paraplegic patients are able to dynamically walk hands-free. \add{To be clear, these tests are not aimed at assessing patient outcomes.}

\add{To be totally transparent, in terms of ``gain tuning'', the local controllers at the joint level will be tuned based on a nominal walking motion and then left fixed. In early stages of optimization, the constraints are adjusted to provide adequate foot clearances given the observed tracking errors in the local joint controllers and small errors in calibration. Post-optimization, a constant bias is sometimes added to a commanded joint profile to compensate for model errors or tracking errors.}

%% file: section_model.tex
\section{The Exoskeleton and Its Dynamic Model}
\label{sec:model}



In this section, we first provide a brief description of the hardware and sensors of the exoskeleton hardware. We next derive a mathematical model of walking for the human-exoskeleton system, which can be represented by a hybrid control system. The model developed here is used later to find periodic walking gaits, develop feedback controllers to stabilize these gaits, and perform numerical simulations of the hybrid control system. The most basic of these generated gaits are evaluated in experiments. 


\subsection{Hardware Description}
\begin{figure}
	\vspace{2mm}
    \includegraphics[width=1\columnwidth]{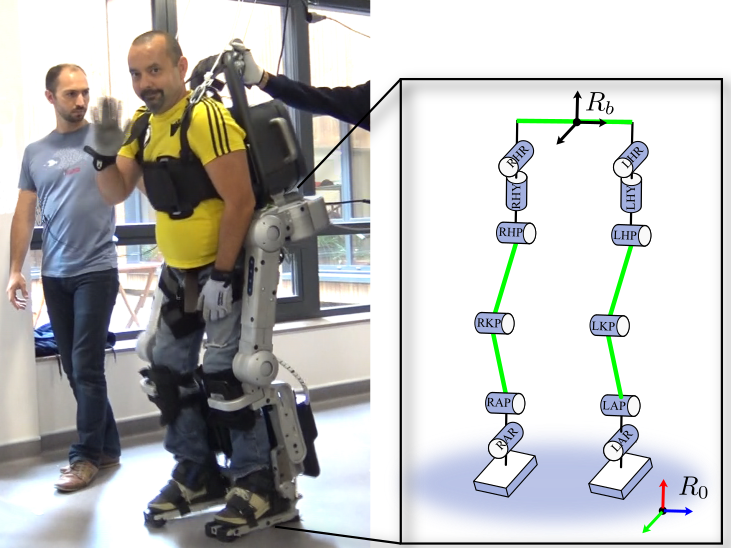} 
    \caption{The kinematic diagram of human-exoskeleton lumped system. The patient is secured to the exoskeleton by means of fasteners located at the ankle, the shin, the thigh, the abdomen, and the torso. The lengths of links highlighted in green are adjustable.}  
	\label{fig:exo-kinematics}
\end{figure}

ATALANTE, developed by the French startup Wandercraft, is a fully-actuated lower-limb exoskeleton intended for use in medical centers for rehabilitation of patients with paraplegia. The exoskeleton consists of 12 actuated joints as shown in \figref{fig:exo-kinematics}: three joints that control the spherical motion of each hip, and in each leg, a single joint for the knee, and two joints for the ankle rotation in the sagittal and frontal plane, respectively.  
Except for the ankle, where a special mechanism is mounted, each degree of freedom is independently actuated by a brushless DC motor. The displacement and velocity of each actuated joint are measured by a digital encoder mounted on the corresponding motor 
as well as by three Inertial Measurement Units (IMUs), with one attached to the torso and one on each leg \add{above the ankles}. Four 3-axis force sensors attached to the bottom of each foot are used to detect ground contact. All electrical components of the exoskeleton are controlled by a central processing unit running a real-time operating system and in charge of high-level computations. 

The mechanical design of ATALANTE allows the leg length and hip width to be manually adjusted to fit to an individual patient's personal measurements. This presents a challenge from the control design perspective in that the controller should be robust to these physical changes of the model. Because the exoskeleton is designed to fully support the user's weight, the user is securely strapped to the device from the feet up to the abdomen as shown in \figref{fig:exo-kinematics}. 


\subsection{Mathematical Representation}

With a goal to study the dynamical behavior of the human-exoskeleton and to avoid over complicating the model by considering the compliant elements present in the human body and exoskeleton linkages, the lumped human-exoskeleton system is modeled as a rigid body system represented by a kinematic tree as shown in \figref{fig:exo-kinematics}. Different from our previous work in \cite{Agrawal2017First}, in which an articulated model of the human torso is considered to allow control of the exoskeleton via the user's upper body posture, in this article, the upper body of the human is modeled as a single rigid link attached to the torso of the exoskeleton. In particular, the patient does not provide any actuation, however, the approximate masses and inertias of the patient are combined in the corresponding links of the exoskeleton. Such a model appears to be appropriate for paraplegic patients who have complete loss of motor input in their lower extremity.

Based on the rigid body assumption, the mathematical representation of the system dynamics can be obtained via the Euler-Lagrangian equations of motion of rigid body dynamics. Specifically, a floating-base generalized coordinate system is considered with the coordinate variables defined as
\begin{align}
  \label{eq:coordinates}
  \Coord = (\BasePos,\BaseRot,\CoordBase) \in \ConfigSpace, 
\end{align}
where $\BasePos \in \RealNum^{3}$ and $\BaseRot \in SO(3)$ denote the relative position and orientation of the exoskeleton's base frame with respect to the world frame respectively, and $\CoordBase \in \RealNum^{12}$ denotes the relative angles of the actuated joints. 
 
This article considers a simplified gait corresponding to flat-footed walking. Specifically, the gait consists of alternating phases of a continuous, single support \emph{swing phase} and an instantaneous, double support \emph{impact phase}, with the stance foot maintained flat on the ground at all times (i.e., the stance foot is not allowed to roll or slip) and the swing foot parallel to the ground at foot strike and foot lift-off. As is common practice in the control design of legged robots \cite{Hurmuzlu2004Modeling, Grizzle2014Models}, the ground contact with the stance foot is considered as non-compliant. Hence, under this assumption, the ground contact can be modeled as a \emph{holonomic constraint}, which enforces the position and orientation of the stance foot to remain constant throughout the swing phase. The dynamical equations of the swing phase with stance foot contact can be obtained as 
\begin{align}\label{eq:swing-dynamics}
D(\q) \ddq + C(\q,\dq)\dq + G(\q) &= B u + J_{st}^T(\q) F_{st},
\end{align}
where $D, C, G$ are the inertia, Coriolis, and gravity matrices respectively that are obtained directly from the exoskeleton's Universal Robot Description File (URDF) \cite{urdfOnline} using FROST \cite{Hereid2017FROST}, an open-source MATLAB toolkit for modeling, trajectory optimization and simulation of hybrid dynamical systems. 
The Jacobian $J_{st}$ of the holonomic constraint
and the ground contact wrench \add{$F_{st} \in \RealNum^{6}$} enforce the holonomic constraint of the the stance foot being flat on the ground \cite{Grizzle2014Models}.

We note in passing that the presented model is a floating-base model. An equivalent pinned-foot model can be developed where the stance foot wrench does not explicitly appear in the dynamic equations. ``\nameref{sidebar:pinned_vs_floating}'' presents the advantages of one over the other. 

Further, under the rigid ground assumption, the swing foot impact with the ground will be considered as plastic (coefficient of restitution is zero) and instantaneous (the impact forces and moments act over an infinitesimal interval of time) impact. During an impact, the coordinate variables of the system remain unchanged. However, the generalized velocities $\dq$ undergo a discrete jump due to the instantaneous change in the generalized momentum. This is captured by a \emph{reset map} $\Delta$ which represents the relationship between the pre-impact states $x^-$ with the post-impact states $x^+$. Let $x = (\q,\dq) \in \ConfigTangentSpace$ be the states of the system dynamics, \add{where $\ConfigTangentSpace$ is the tangent space of $\ConfigSpace$}, the hybrid system model of the flat-footed walking of the exoskeleton can be written as
\begin{equation} \label{eq:hbd_model}
  \Sigma :  
  \begin{cases}
      \dot{x} =  f(x) + g(x)u,& x \notin S,\\
       x^+ = \Delta(x^-),     & x\in S.
  \end{cases}
\end{equation}
where $S$ is the guard or switching surface which determines the specific condition (i.e., the swing foot impacting the ground) that triggers the discrete events and the vector fields $f, g$ are from the continuous-time swing-phase dynamics in \eqref{eq:swing-dynamics}. See \cite{Grizzle2014Models} for a thorough discussion on obtaining dynamical models for bipedal mechanical systems.

%% file: section_ghzd_controller_overview.tex
\section{Gait Objectives, Important Constraints, and Controller Architecture}
\label{sec:controller_overview}

The embedded control system needs to generate comfortable, robustly stable walking gaits that respect mechanical limits of the exoskeleton, such as joint and torque limits,  initiate smooth (not jarring) foot contact with the ground, and satisfy ground contact constraints that avoid slipping. These requirements will be the main focus of the paper. In previous work \cite{Agrawal2017First}, we provided additional features in the closed-loop system that may provide an intuitive means for the user to regulate walking speed, and eventually, direction.    


\subsection[Gait Design Objectives]{Gait Design Objectives}
\label{sec:GaitDesignObjectives}

The gaits designed in this paper are destined for testing in a medical facility where an engineer or therapist will provide external commands for walking speed. A wearer's directional-and-speed-control interface will be tested at a later stage. Gaits will be designed here for walking in a straight line at speeds that vary from $-0.3$ m/s to $0.3$ m/s. For comparison purposes, the relaxed human walking gait is approximately $0.9$ m/s.

The time duration of a step will be set to $0.7$~s. Our observation is that shorter step times closer to $0.5$~s, while easier to stabilize, are uncomfortable for the user. To limit the transmission of vibrations from the exoskeleton to the user, the impact of the swing foot with the ground needs to be carefully regulated. Just before impact, we try to achieve near-zero forward and lateral velocities of the foot with respect to the ground, while the downward velocity of the foot is between $-0.3$ and $-0.1$~m/s. The upper bound ensures a transversal intersection of the foot with the ground, a key guard condition in the hybrid controller, while the lower bound is for user comfort.  

It is desirable for the user to be able to maintain an upright posture when using the exoskeleton to limit demands on abdominal and dorsal muscles, which may have been weakened through prolonged use of a wheelchair. We have settled on left-right swaying motions that are less than two degrees, and a forward lean angle that is between two and six degrees. For user safety, the knee angles are bounded above five degrees away from straight and the ankles are limited to $\pm 23$ degrees. \add{Joint safety limits are imposed through a combination of hardware limits and software enforced limits.}
 
\subsection{Ground Contact} 
\begin{figure}
	\begin{centering}
        \vspace{2mm}
		\includegraphics[width=1.0\columnwidth]{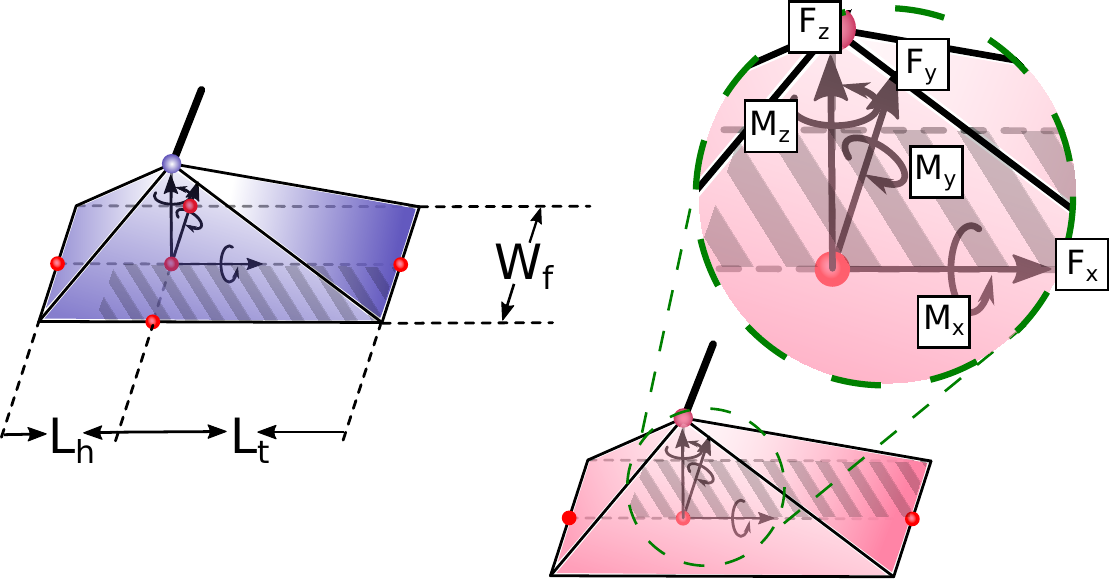}
	\end{centering}
	\caption{Depicts feet of the exoskeleton, with blue the left foot pink the right foot. \add{$L_h$ and $L_t$ denote the distance from the point of reference to the heel and toe respectively. $W_f$ denotes the width of the foot.} Shaded area depicts the desired location of the \add{Zero Moment Point (ZMP)}. $[F_x,F_y,F_z,M_x,M_y,M_z]$ is the resultant ground reaction force and moment acting on the foot during support\add{, described in the body coordinate of the foot}.}
    \label{fig:feet_zmp}
\end{figure}

The holonomic constraints for modeling ground are taken from \cite{Grizzle2014Models}. The key things to note are that the ground cannot pull on a foot and a ``friction cone'' must be respected to avoid foot slippage, namely,
\begin{align}
	F_z > 0, \\
    F_x^2 + F_y^2 \leq \mu^2 F_z^2, 
\end{align}
where, $[F_x,F_y,F_z]$ is a vector of ground reaction forces acting on the stance foot, as shown in \figref{fig:feet_zmp}, and $0 < \mu < 1$ is the coefficient of friction. 

For the gaits used in this study, we will simplify the motions of the exoskeleton by imposing that the stance foot remains flat on the ground. As explained in \cite{Grizzle2014Models}, this requires moment constraints so that that foot does not roll about one of its axes. The stance width is set at $27$~cm. This relatively wide stance limits rolling about the outer edge of the stance foot, which is typically harder to recover from than rolling inwards on the stance foot, while also promoting lateral stability. To provide additional robustness against foot rotation, we design the gaits so that the Zero Moment Point (ZMP) \cite{vukobratovic2004zero} lies in the shaded areas shown in \figref{fig:feet_zmp}; though it is not exactly the same thing, for the purpose of this paper, the reader can think of the ZMP as being the Center of Pressure (CoP) of the forces distributed on the sole of the foot. In addition, due to the relatively heavy battery pack mounted just behind the user's hips, the center of mass of the exoskeleton is towards the heel of the foot when the leg is straight. Designing gaits with the ZMP towards the forward section of the foot prevents the exoskeleton from rolling backward on its stance foot. In our experience, rolling forward on the foot has not been a concern.

\subsection{Other Objectives}

To reduce the possibility of the swing foot contacting the ground prematurely, gaits are designed with relatively large foot clearance in the middle of the step. The heel is designed to be $10$~cm above the ground and the toe to be $5$~cm above the ground. Larger foot clearance results in easier handling of terrain irregularities and trip recovery via foot placement. Potential downsides include greater torque requirements and motions that are closer to joint limits. Because flat walking is assumed, gaits need to be designed such that the feet are parallel to the ground during lift off and impact. 

\subsection{Controller Architecture}
\begin{figure*}
	\centering
	\includegraphics[width=0.8\linewidth]{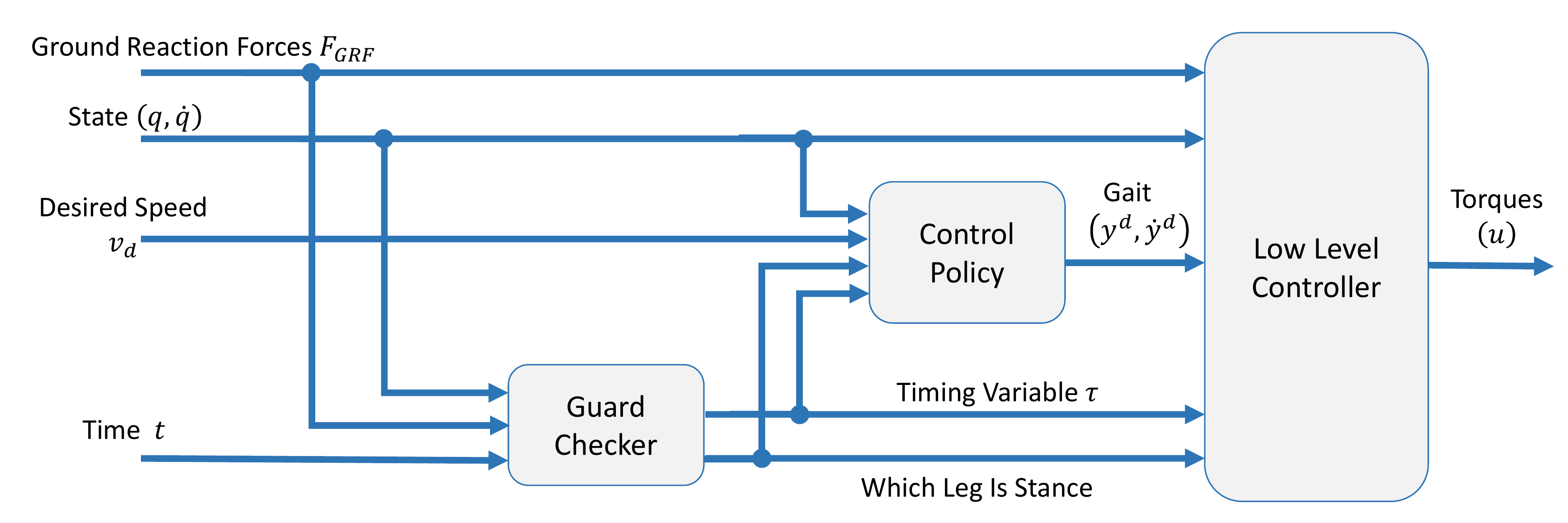}
	\caption{Overview of the controller structure. The Guard Checker handles detecting when to swap legs. The Control Policy specifies the desired gait and the means to achieve it. The Low-level Controller translates control policy commands to desired joint-level trajectories and achieves them via PD control.}
    \label{fig:controller_overview}
\end{figure*}

The overall control structure is shown in \figref{fig:controller_overview}. The  Control Policy is responsible for specifying the evolution of key quantities of the exoskeleton, such as torso angle, swing leg angle (imagine a line from the hip to the ankle), and stance leg length (once again, a line from the hip to the ankle). These synthetic quantities are often more intuitive for the control engineer and the test engineer to use when specifying and discussing gait designs. 

The Low-level Controller is responsible for associating the synthetic high-level quantities to the individual actuators of the exoskeleton. \strike{When push comes to shove} During early testing, the simpler the low-level controller, the easier it is to make rapid changes and uncover bugs.
\add{The low-level joint controllers assure trajectory tracking with less than two degrees of error.}
The main task to be discussed later in the paper is \add{therefore} the association of high-level control policy objectives to individual (or pairs of) joints and motors.

As its name suggests, the Guard Checker monitors quantities associated with events in a gait. In the real world, leg swapping is a control decision and not a discrete, capturable event as it would appear in an ideal simulator. The guard event for leg swapping is defined here in terms of step duration and measured vertical ground reaction force. In general, a gait timing variable is a strictly monotonically increasing quantity that varies from $0$ to $1$ over the course of a step. Here we use $\tau=\frac{t-t_0}{T_p}$, where $t_0$ is the starting time of the current step and $T_p$ is the time duration of a step. The guard for leg swapping is then
\begin{equation}
\begin{aligned}
	(GRF^z_{swing} > GRF_{min} \text{~and~} \tau > 0.5) \\
    \text{~or~} \tau > 1,
\end{aligned}
\end{equation}
where $GRF^z_{swing}$ is the measured vertical component of the ground reaction force acting on the swing foot and $GRF_{min}$ is a chosen minimum threshold.

%% file: section_phzd.tex
\section{PHZD: Control Policy Design on the Basis of a Single Periodic Gait}

Given the hybrid model of the system as in \eqref{eq:hbd_model}, the first objective of this article is to design a feedback control policy that creates and robustly stabilizes a single periodic solution of the exoskeleton. Specifically, we view the combined fully actuated exoskeleton and its user as a 3D bipedal mechanism. In this section, a brief description of the well-studied virtual constraints-based feedback control law for a single periodic gait is presented. In a later section, a control methodology that addresses more complex dynamical behaviors of the exoskeleton is introduced.

\subsection{Virtual Constraints} 
At the core of this method is the design of a set of virtual constraints that modulate the joint trajectories of the system in order to achieve certain desired behaviors \cite{ames2014human,Westervelt2007Feedback}. Enforcing virtual constraints results in a lower dimensional representation of the full-order system---termed the partial hybrid zero dynamics (PHZD)---that captures the natural dynamics of the mechanical system. While the PHZD is a reduced-order model, it does not involve any approximations of the dynamics. Solutions of the PHZD are solutions of the original system model under feedback control.

The virtual constraints are defined as the difference between actual physical quantities and their desired evolution, and then posed as outputs of the system that are to be zeroed by a feedback controller. In general, the actual outputs, $y^a$,  represent important kinematic functions of the robot: they could be as simple as particular joint variables, such as the hip and knee angles, or they could also be more complicated functions of robot states, such as swing foot orientation in the world frame or forward velocity of the pelvis. The desired outputs are often represented by a group of parametrized curves with a timing variable. In their traditional form, the virtual constraints are synchronized through a state-based timing variable. Work in  \cite{WANGChevallereau2011} and \cite{ReCoHeHuAm16} show that such a state-dependent design is not strictly required. In order to present the main idea of a virtual-constraints-based feedback control design, a state-based phase variable is still assumed in this section.


For the particular case of the human-exoskeleton system with powered ankle joints, the actual outputs are chosen to be a combination of a velocity regulating term $y_1^a$ and posture modulating terms $y_2^a$. Specifically, the velocity regulating output is the forward hip velocity of the exoskeleton, and posture modulating outputs are chosen to represent the synchronized motion of the remaining actuated joints. Hence, the virtual constraints for the exoskeleton are defined as
\begin{align}
\label{eq:y1}
  y_{1}(\q,\dq,\alpha) &= y^a_{1}(\q,\dq) - y^d_{1}(\alpha), \\
\label{eq:y2}
  y_{2}(\q,\alpha) &= y^a_{2}(\q) -
                     y^d_{2}(\theta(t),\alpha),
\end{align}
where $y_{1}$ and $y_{2}$ are relative degree 1 and (vector) relative degree 2 by construction, \add{and $\theta$ is a ``phasing variable'' \cite{kolathaya2018input}}. To fully determine a motion of the entire system, the outputs, $(y^a_{1}, y^a_{2})$, must be linearly independent and their rank must be equal to the number of actuators in the system. $y^a_{1}$ is set to be the forward hip velocity, while $y^a_{2}$ is set to be the all joint angles of the exoskeleton except for the sagittal stance ankle \cite{Gurriet2018Towards}.


\begin{figure}
  \centering
  \subcaptionbox[]{Periodic Orbit
    \label{fig:periodic-orbit}
  }{
    \includegraphics[width=0.33\columnwidth]{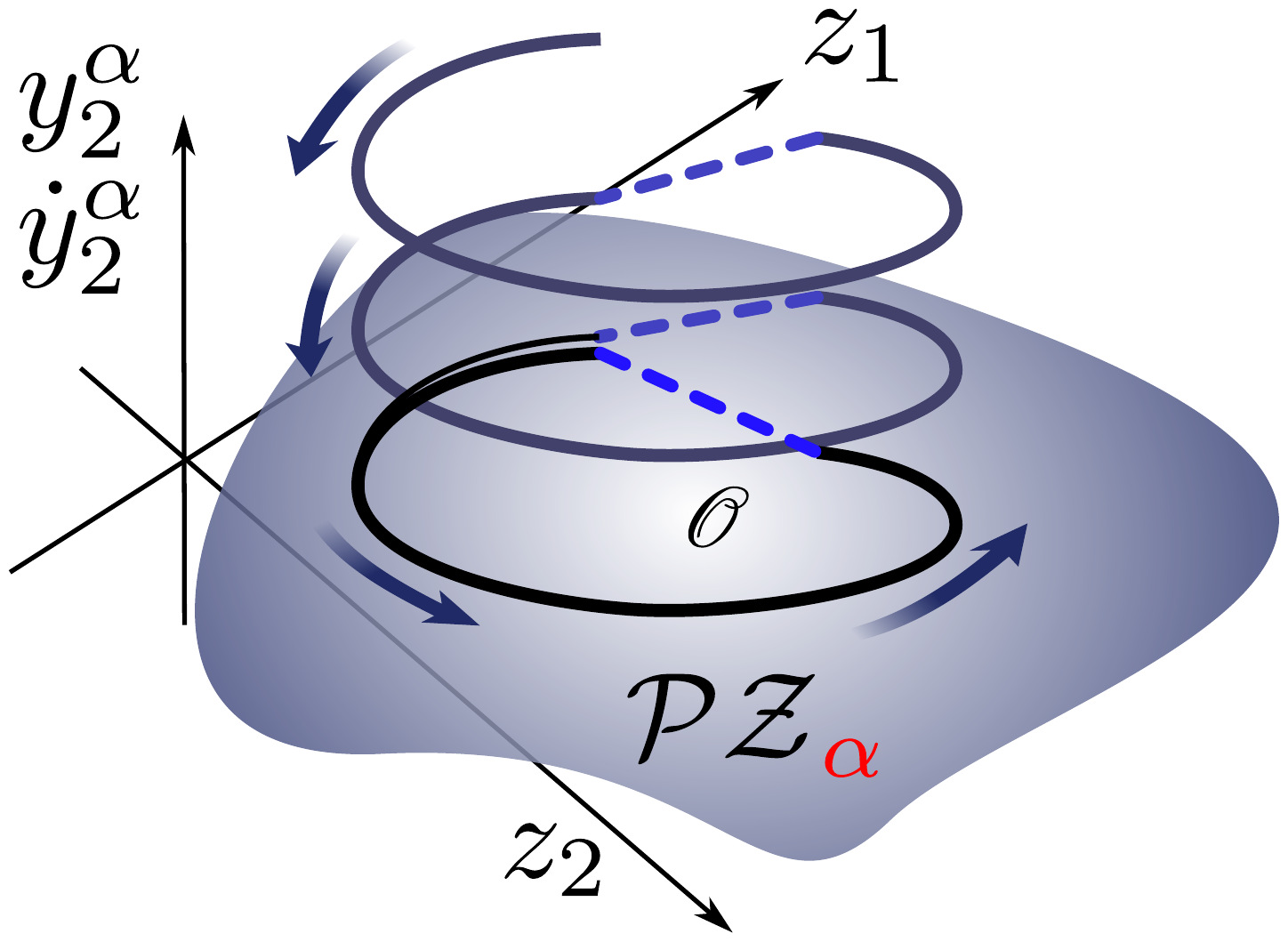}    
  }
  \subcaptionbox[]{Hybrid Invariance
    \label{fig:hybrid-invariance}
  }{
    \includegraphics[width=0.57\columnwidth]{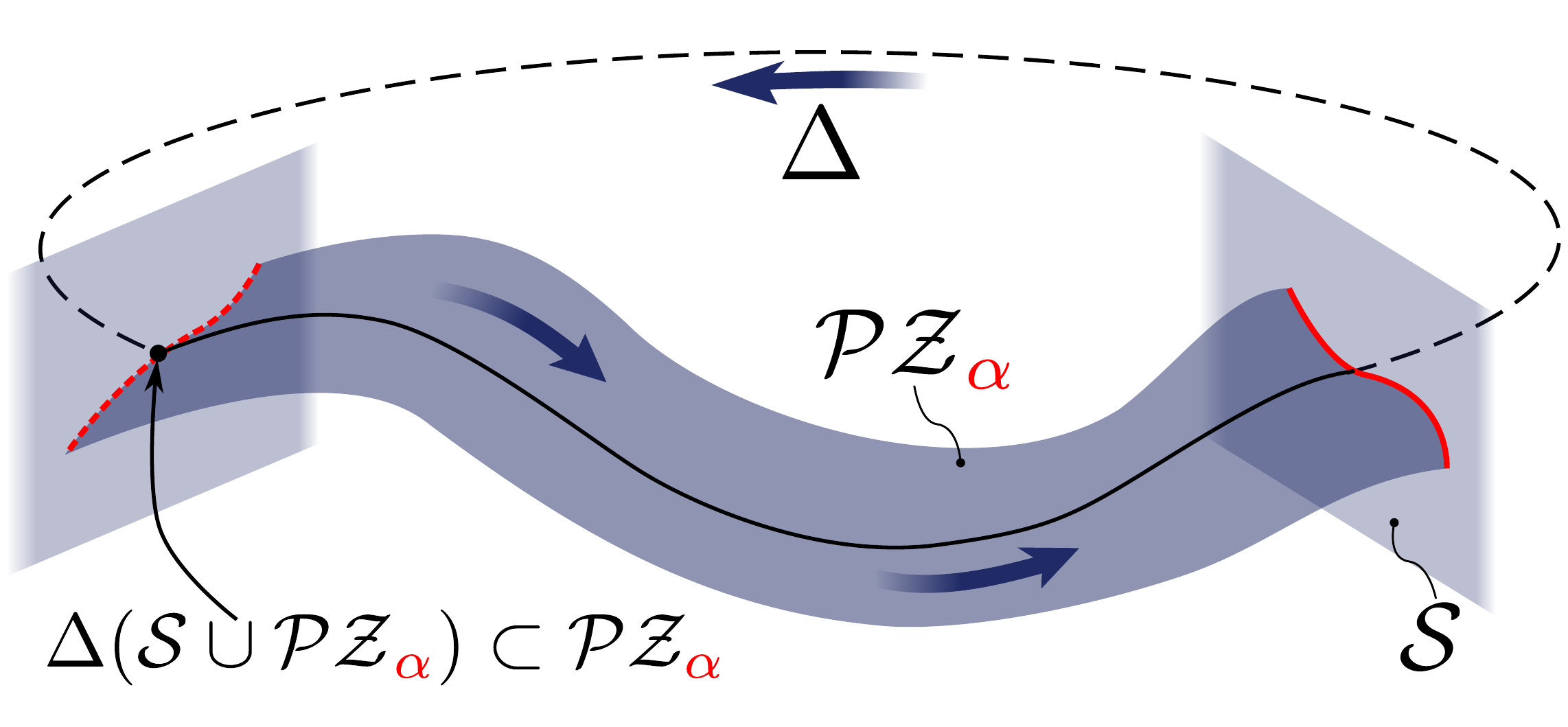}    
  }
  \caption{Illustration of a periodic orbit on the partial hybrid zero dynamics (PHZD) surface. A PHZD surface admitting a periodic orbit is designed by selecting a proper parameter set $\alpha$  in the virtual constraints through optimization.}
\end{figure}

\subsection{Input Output Linearization.} 

With the goal of driving the virtual
constraints in \eqref{eq:y1} and \eqref{eq:y2} to zero exponentially, the following feedback control law, based on Input Output Linearization, is considered,
\begin{align}
  \label{eq:io-control}
  u = -\decouplingmatrix^{-1} \left( \left[\!\!
                     \begin{array}{c}
                       L_{f} y_{1}(\q,\dq,\alpha) \\
                       L_{f}^2  y_{2}(\q,\dq,\alpha)
                     \end{array} \!\! \right] \right.    & +
                     \left[\!\!
                     \begin{array}{c}
                       \epsilon y_{1}(\q,\dq,\alpha) \\
                       2 \epsilon \dot{y}_{2}(\q,\dq,\alpha)
                     \end{array} \!\! \right] \nonumber \\
  & \left. + \left[\!\!
  \begin{array}{c}
    0 \\
    \epsilon^2  y_{2}(\q,\alpha)
  \end{array} \!\! \right] \right),
\end{align}
with a control gain $\epsilon > 0$ and \emph{decoupling matrix}
\begin{align}
  \decouplingmatrix = \left[\!\!
  \begin{array}{c}
    L_{g} y^a_{1}(\q,\dq) \\
    L_{g}L_{f} y_{2}(\q,\dq,\alpha)
  \end{array} \!\! \right],
\end{align}
where $L_f$ and $L_g$ represents the Lie derivatives with respect to the vector fields $f(x)$ and $g(x)$ in \eqref{eq:hbd_model}. With a specific choice of virtual constraints, the decoupling matrix $\decouplingmatrix$ is invertible. Applying this control law to \eqref{eq:hbd_model} yields linear output dynamics of the form:
\begin{align}
  \label{eq:output-dynamics}
  \dot{y}_{1} &= -\frac{1}{\epsilon} y_{1}, \\
  \label{eq:output-dynamics-2}
  \ddot{y}_{2} &= -2\frac{1}{\epsilon} \dot{y}_{2} -\frac{1}{\epsilon^2}
                 y_{2},
\end{align}
which has an exponentially stable equilibrium at the origin. 

Hence, the directly actuated variables of the system are regulated to a reduced-dimensional surface called the \emph{zero dynamics} that is invariant within the duration of continuous swing phase, \add{as illustrated in \figref{fig:periodic-orbit}} \cite{Westervelt2007Feedback,ames2014human}. Yet, due to the discrete joint velocity changes in the system's states at swing foot impact, the controller in \eqref{eq:io-control} does not necessarily guarantee the reduced-dimensional surface is invariant through the impact. It is shown in \cite{Westervelt2007Feedback} that, if there exists a set of virtual constraints such that the reduced-dimensional zero dynamics surface is invariant through impact, then the full-order dynamics of the hybrid system model restricts to a \emph{hybrid-invariant} reduced-dimensional submanifold. The restriction dynamics and invariant surface is the \emph{hybrid zero dynamics} (HZD). This requires one to find a set of parameters $\alpha$ for the virtual constraints such that the zero dynamics is invariant through impact maps \add{(see \figref{fig:hybrid-invariance})}. Finding such parameters is typically formulated as a nonlinear optimization problem \cite{ames2014human, Hereid2018Dynamic}. The advantage of studying the hybrid zero dynamics manifold is that the evaluation of orbital stability of the full-order system can be performed on the reduced-dimensional zero dynamics. 

\subsection{Generation of a Periodic Gait}

Periodic walking gaits are periodic orbits of the corresponding hybrid system model. 
A solution $\varphi(t)$ of the hybrid system in \eqref{eq:hbd_model} is \emph{periodic} if there exist a finite $T>0$ such that $\varphi(t+T) = \varphi(t)$ for all $t \in [t_0,\infty)$. A set $\Orbit \subset \ConfigTangentSpace$ is a periodic orbit of the system if $\Orbit = \{\varphi(t) | t \geq t_0\}$ for some periodic solution $\varphi(t)$. The stability of the periodic orbit can be determined by the stability of the fixed point by evaluating the spectral radius of the Jacobian of the Poincar\'{e} map at the fixed point. More specifically, if all eigenvalues lie within the unit circle, i.e., have magnitude less than $1$, then the periodic orbit is locally exponentially stable.

To design a periodic gait for the hybrid system model of the exoskeleton, a direct collocation-based gait optimizer is used.
\add{The mathematical foundation behind the optimization technique used is briefly introduced in the sidebar ``\nameref{sidbar:direct_collocation}''}.
Other ways of solving the optimization can be used, such as single shooting methods, but direct collocation was found to be the fastest and most efficient way to solve this problem \cite{Hereid2018Dynamic}; stable walking gaits are obtained in minutes. Considering that our goal in this section is to find parameters for the virtual constraints instead of open-loop trajectories, we incorporate the feedback controller into the optimization in a way that is similar to holonomic constraints. Instead of enforcing the control input directly as in \eqref{eq:io-control}, we impose equality constraints on system states such that they satisfy the output dynamics in \eqref{eq:output-dynamics} and \eqref{eq:output-dynamics-2}.
Furthermore, the hybrid-invariance is enforced in the periodic gait optimization as a constraint.
For a detailed setup of the optimization problem in the context of the exoskeleton, the reader is referred to \cite{Agrawal2017First, Gurriet2018Towards}.
Torque limits and joint position and velocity limits of the ATALANTE mechanism are directly enforced as boundary conditions on decision variables in the optimization, whereas friction cone and zero moment constraints of foot contacts are enforced as extra physical constraints.
In addition, several constraints are considered in the optimization in order to narrow down the search space and address certain aspects specific to human-friendly walking. Impact velocities, ZMP position, CoM position and torso orientation are examples of the many constraints that need to be considered.

The result of the optimization is a single periodic orbit and a feedback controller that renders it locally exponentially stable in the model of the user plus exoskeleton. Simulations of the controller can be found in \cite{Agrawal2017First} and are not given here. The experiments reported later are based on the design process described above.

%% file: section_optim_plus_machinelearning.tex
\section{G-HZD: Harnessing the Power of Modern Optimization Techniques}
\label{sec:OptimPlusML}

When the HZD and PHZD methods were created, it would take several hours for the computation of a single periodic orbit. At that time, it was very important to be able to build a controller for rendering the periodic orbit locally exponentially stable directly from the orbit itself, without requiring further optimizations. The situation today is totally different as shown in Table \ref{table:comp-results-single}. So, how to harness this power?

\begin{table}
  \centering
  \caption{Performance comparison between direct-collocation optimization vs
    classic shooting approaches on a 5-link planar biped (single-shooting) and a
    7-link spring-leg planar biped, respectively.}
  \begin{tabular}{@{\extracolsep{\fill}}cc}
    \hline
    Method & CPU time (s) \\
    \hline
    Single shooting (\texttt{fmincon}) &  162.59  \\
    Direct collocation (\texttt{IPOPT}) & 1.60  \\
    \hline
  \end{tabular}
  \label{table:comp-results-single}
  \quad
  \begin{tabular}{@{\extracolsep{\fill}}cc}
    \hline
    Method & CPU time (s)  \\
    \hline
    Multiple shooting (\texttt{fmincon}) &  5027.45  \\
    Direct collocation (\texttt{IPOPT}) & 41.47  \\
    \hline
  \end{tabular}
  \label{table:comp-results}
\end{table}

The control loops on the exoskeleton run at 1 KHz and the duration of a walking step is on the order of 500 ms to 750 ms. Hence, online Model Predictive Control (MPC) (i.e., iteratively solving in real time a finite-horizon constrained optimization problem) is simply not possible for models with twenty or more state variables. Explicit MPC is not possible either because one would have to numerically sample the state space, do the optimization offline, and then store the control actions for use online. A sparse uniform grid of ten samples per dimension would require $10^{20}$ optimizations; random sampling will provide a smaller, more effective discretization of the state space, but not enough is gained to handle $n\ge 20$. So, what to do?

Reference \cite{IJRR_DaGrizzle2017} introduces G-HZD, Generalized Hybrid Zero Dynamics, a new approach to control design for a class of high-dimensional nonlinear systems. As with PHZD, the design process for G-HZD begins with trajectory optimization to design an open-loop periodic walking motion of the high-dimensional model. It differs from PHZD in that it exploits the fact the trajectory optimization can be done rapidly to add to this periodic solution a carefully selected set of additional open-loop trajectories of the model that steer toward the nominal motion, thereby directly building in stability. A drawback of trajectories is that they provide little information on how to respond to a disturbance. To address this shortcoming, Supervised Machine Learning is used to extract a low-dimensional state-variable realization of the open-loop trajectories. The periodic orbit is now an attractor of the low-dimensional state-variable model but is not attractive in the full-order system. The special structure of mechanical models associated with bipedal robots is used to embed the low-dimensional model in the original model in such a manner that the desired walking motions are locally exponentially stable. 

In the following, we give the main ideas underlying G-HZD for models given by ordinary differential equations; the reader is referred to  \cite{IJRR_DaGrizzle2017} for the small technical changes required to deal with hybrid models.


\begin{figure}
	\centering
	\includegraphics[width=0.8\columnwidth]{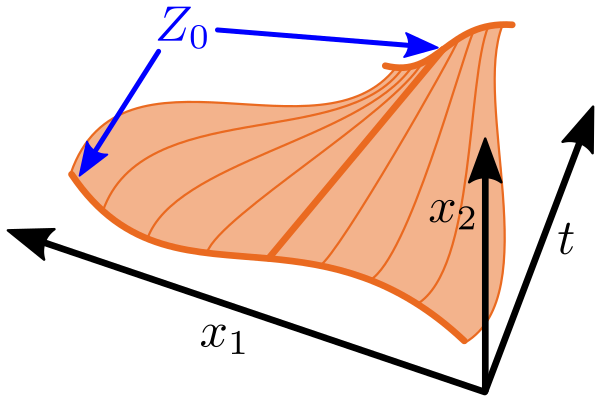}
   	\caption{A collection of trajectories forming a smooth surface; while this is the desired outcome, it is not guaranteed. Direct-collocation-based optimization is used to design the trajectories. }
    \label{fig:MLstep1}
\end{figure}


\subsection{Step 1: Constructing $Z_0$, the boundary of the generalized HZD surface, from a collection of periodic orbits.}
The \add{dynamic} model of the exoskeleton is decomposed into a low-dimensional weakly actuated portion corresponding roughly to the $x-y$-coordinates and velocities of the hips \add{(in the world frame)} and a strongly actuated portion of the model that captures the hips, knees, and swing ankle joints. Specifically, the \add{dynamic} portion of the hybrid model \eqref{eq:hbd_model} is decomposed as
\begin{equation}
\label{eq:Decomposed Model}
\begin{aligned}
\dot{x}_1&= f_1(x_1,x_2,u_1)~~~x_1\in \real^{n_1}\\
\dot{x}_2&= f_2(x_1,x_2,u_1,u_2)~~~x_2\in \real^{n_2},
\end{aligned}
\end{equation}
where $x_1$ represents the ``weakly actuated'' portion of the model, $u_1$ are the stance ankle torques, $x_2$ captures the strongly actuated part of the model and the remaining actuators $u_2$. With this decomposition, $n_1 \ll n_2$; indeed, for the exoskeleton as modeled here in single support, $n_1 = 4$ and $n_2 = 20$.

Next, a library of gaits is constructed by uniformly discretizing a bounded set of initial conditions for $x_1$.  Without loss of generality, each periodic gait is assumed to start from the origin, i.e., hip positions $(x; y)$ are at the origin.  As a consequence, the periodic gaits are parametrized by the $(x; y)$-velocities of the hip, corresponding to walking forwards, backwards, and sideways. Turning is not addressed presently but will be in the near future.

A nonlinear mapping is then constructed between the $x_1$ and $x_2$ states for the periodic orbits such that $x_2 = \gamma(x_1)$. The function $\gamma$ is called an insertion map and can be constructed in several ways.  One effective approach is to numerically fit a surface to the initial conditions of the periodic orbits through machine learning tools, resulting in
\begin{equation}
\label{eq:Z0}
Z_0:= \{ x=:(x_1; x_2) | x_2 = \gamma(x_1) \},
\end{equation}
as illustrated in \figref{fig:MLstep1}. In the ideal case, every point on the surface $Z_0$ corresponds to an initial condition for a periodic orbit. 


\subsection{Step 2: Constructing $Z$, the Generalized HZD Surface.}

Assume now that one of the periodic orbits has been selected, corresponding to a point $\xi^* \in Z_0$. The process of stabilizing this particular periodic orbit is based on computing a judiciously selected set of open-loop trajectories of the full-order model that ``approach'' the periodic solution. Specifically, let $B$ be an open ball about $\xi^*$ and select a ``contraction factor'' $0 < c < 1$.  For each initial condition $\xi \in B \cap Z_0$, a solution of the ODE \eqref{eq:Decomposed Model} with initial condition $\xi$  is sought that satisfies the physical constraints given in Section \nameref{sec:GaitDesignObjectives}, terminates in $Z_0$ at time $t=T_p$, and approaches the periodic solution as measured by $||\varphi_\xi(T_p) - \xi^*|| \le c ||\xi - \xi^*||$; \add{transient solutions that approach periodic solutions are called transition gaits.}

Denote the corresponding input and state trajectories parametrized by the initial condition $\xi \in B \cap Z_0$ as 
\begin{equation}
  \label{eq:PeriodicBehavior}
  \begin{aligned}
  u_\xi:[0, T_p]& \to \real^m, \\
  \varphi_\xi:[0, T_p]& \to \real^n.
  \end{aligned}
\end{equation}
If all has gone well, the process has resulted in a smooth surface $Z$ of dimension equal to the dimension of $Z_0$ plus one (due to time), as shown in \figref{fig:MLstep1}.
As discussed in \cite{IJRR_DaGrizzle2017}, such solutions may or may not exist and in principle could be very ``ugly'' functions of $\xi$ (i.e., non-smooth, etc.).
We note that because for all $\xi \in Z_0$, $\varphi_{\xi}(T_p) \in Z_0$, trajectories that start in $Z_0$ can be continued indefinitely.

\subsection{Step 3: State-variable Realization of the Open-loop Trajectories Forming $Z$.}
While the above process results in complete trajectories for the model, we do not directly use these for tracking because: (a) If the system starts on the surface $Z$, and a perturbation occurs between $t=0$ and $t=T_p$, it would attempt to converge back to the potentially-far-away original-trajectory $\varphi_xi(t)$ until $t=T_p$, only at which point would a trajectory-based controller update the desired trajectory; and (b), if the system does not start on the surface, there is no obvious choice of a trajectory to follow. 

We first address (a) by seeking a means to continuously update the desired evolution of the system so as to immediately respond to a perturbation.  In fact, we will design a low-dimensional differential equation that evolves on the surface and has the desired periodic orbit as its locally stable and attractive limit cycle. A solution to (b) will be given in Step 5.

\begin{figure}
  \centering
  \subcaptionbox[]{Disturbance occurs at time $t_d$    
  }{
    \includegraphics[width=0.45\columnwidth]{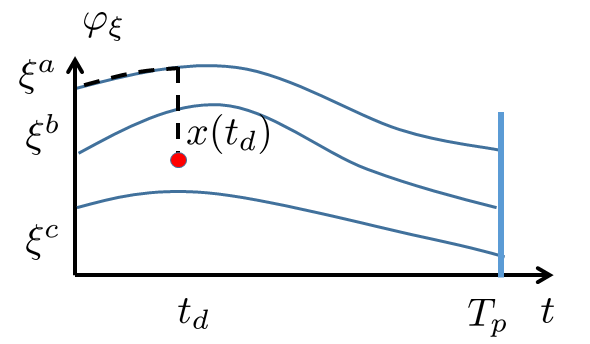}
  }
  \subcaptionbox[]{Proposed ``course'' correction    
  }{
    \includegraphics[width=0.45\columnwidth]{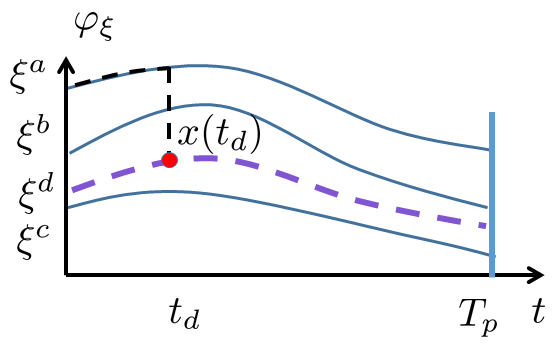}
  }
  \caption{Gedanken Experiment. (a) Assume the system is initialized at $\xi^a$, with input, $u_{\xi^a}(t)$, being applied, and hence its solution is evolving along $\varphi_{\xi^a}(t)$. Suppose that at time $t_d$, an ``impulsive'' disturbance instantaneously displaces the solution to a point $x(t_d)$. What input should be applied? (b)  If there exists an $\xi^d \in Z_0$ such that $x(t_d)=\varphi_{\xi^d}(t_d)$, then applying the input $u_{\xi^d}(t)$ for $t_d \le t <T_p$ will move the system toward the equilibrium in the sense that $||\varphi_{\xi^d}(T_p)) -\xi^*|| \le c ||\xi^d-\xi^*||$. This thought process leads to the ``interpolation'' or ``learning'' conditions in \eqref{eq:MagicConditions}.}
 \label{fig:Gerdunken}
\end{figure}

\figref{fig:Gerdunken} motivates a condition for ``automatically re-planning'' so as to respond to a disturbance, namely
\begin{equation}
\label{eq:InterpConditions}
x(t)=\varphi_{\xi}(t) \implies u(t,x(t))=u_\xi(t).
\end{equation}
This is an implicit interpolation condition for specifying a control response at each point of the surface $Z$ in \figref{fig:MLstep1}.
As explained further in \cite{IJRR_DaGrizzle2017}, a solution to \eqref{eq:InterpConditions} can be constructed if one can find two $T_p$-periodic functions, $\nu$ and $\mu$, satisfying the following conditions on the trajectory data: for all $0 \le t < T_p$
\begin{equation}
\label{eq:MagicConditions}
\begin{aligned}
\nu(t,\pi_1 \circ \varphi_{\xi}(t))&=\pi_2 \circ \varphi_{\xi}(t)\\
\mu(t,\pi_1 \circ \varphi_{\xi}(t))&=u_\xi(t),
\end{aligned}
\end{equation}
where $\pi_i: \real^n \to \real^{n_i}$ are the canonical projections (i.e., $\pi_1(x_1,x_2)=x_1$ and $\pi_2(x_1,x_2)=x_2$). Moreover, it can then be shown 
that (i) the trajectories of \figref{fig:MLstep1} are solutions of the reduced-order model
\begin{equation}
  \label{eq:ZeroDynamics}
  \begin{aligned}
  \dot{x}_1&= f_1(x_1,\nu(t, x_1),\mu_1(t,x_1)),\\
  x_2 &= \nu(t, x_1),
  \end{aligned}
\end{equation}
and that (ii) the periodic orbit $\varphi_{\xi^*}:[0, T_p) \to \real^n$ is locally exponentially stable. In fact, by construction, all initial conditions $\xi \in Z_0$ for which feasible model solutions have been found result
in trajectories that converge to the periodic orbit; hence, as one builds the feasible solutions in \eqref{eq:PeriodicBehavior}, one is constructing the domain of attraction of the periodic orbit in the G-HZD surface, $Z$. 

\begin{figure}
	\centering
    \includegraphics[width=0.9\columnwidth]{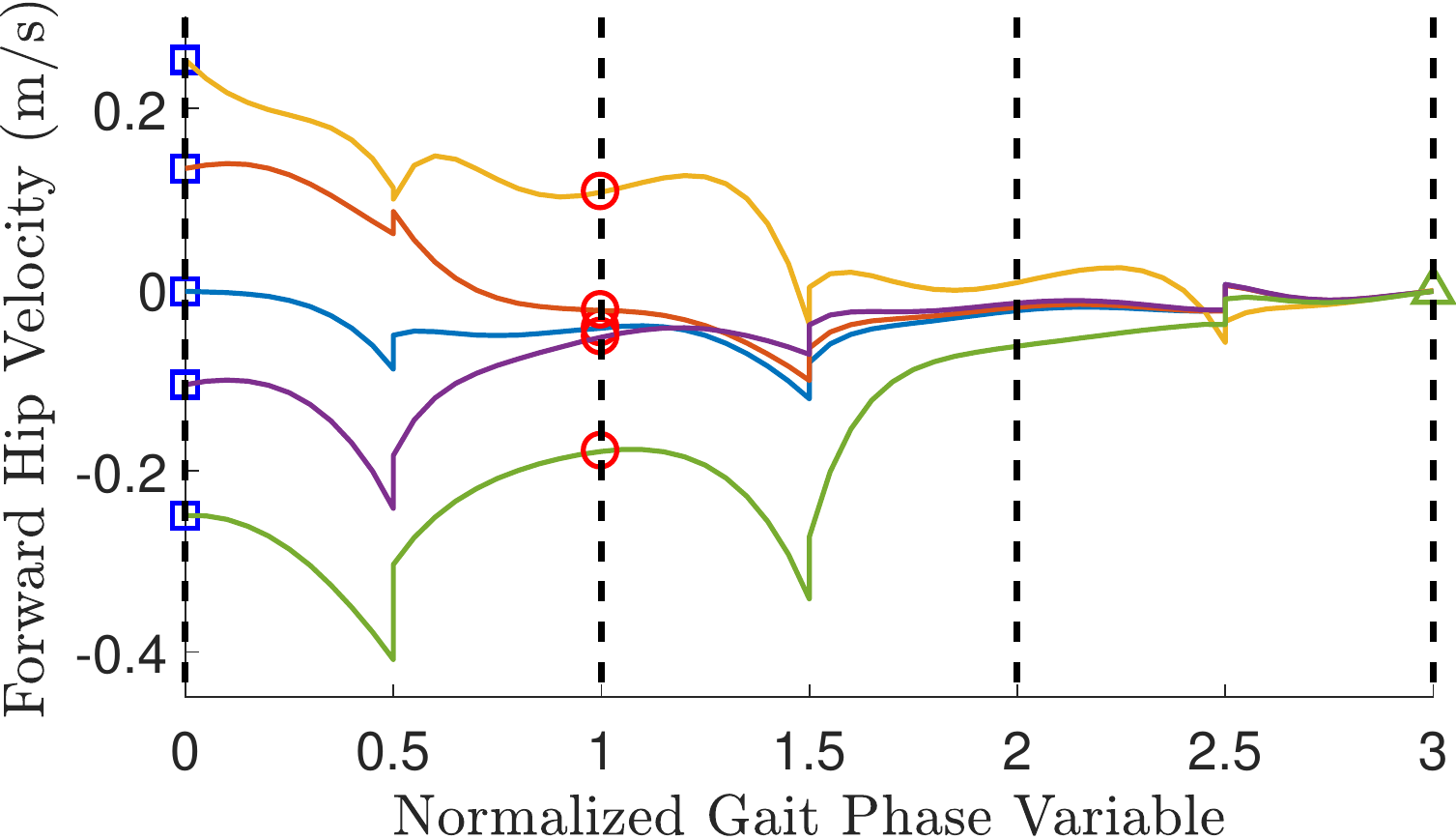}
	\caption{Shows forward velocity of the hip resulting from the transition optimizations. \strike{Black dashed line marks the end of each step in the optimization.} \add{Black vertical dashed lines mark the midpoint of each (robot) step in the optimization, whereas the discontinuities (jumps) in velocity between two dashed lines arise from the rigid impact at the end of each step.} Squares, circles, and the triangle mark points where the state of the system is in $Z_0$. The blue squares mark starting states obtained from periodic gaits while the green triangle marks the ending desired state also obtained from a periodic gait. The red circles were made to be in $Z_0$ through an optimization constraint. The gait trajectory from the blue square to the red circle is what is used as training data for the Supervised Machine Learning.}
    \label{fig:velocity_transition_opt}
\end{figure}

\figref{fig:velocity_transition_opt} shows the sagittal-plane hip velocity for several initial conditions in $Z_0$. Two things are important to note: (a) the convergence to the nominal orbit; and (b), the trajectories are feasible solutions of the full-order model. 

Next we deal with how to find functions satisfying the conditions in \eqref{eq:MagicConditions}.

\subsection{Step 4: Supervised Machine Learning to Extract $\nu$ and $\mu$ from Optimization Data}

\begin{figure}
	\centering
	\includegraphics[width=0.8\columnwidth]{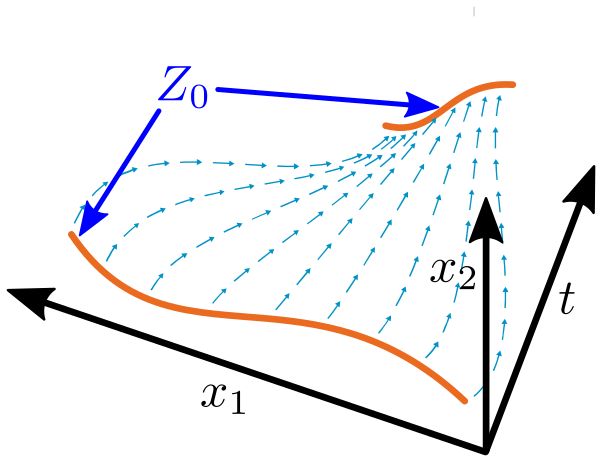}
   	\caption{A vector field is constructed that gives rise to the trajectories, so it is a state-variable realization. Supervised Machine Learning (fancy regression) and model structure are used to ``extract'' the vector field from the trajectory optimization data.}
    \label{fig:MLstep2}
\end{figure}

\add{The importance of finding a differential equation (i.e., vector field) realization of the trajectories in Figure~\ref{fig:MLstep1} is that the differential equation is an automatic, instantaneous re-planner of the system's evolution when it is perturbed off a nominal motion. Figure~\ref{fig:MLstep2} shows such a realization. 
Solving for the functions in \eqref{eq:MagicConditions}  is the key to constructing the vector field. Doing so analytically would be a challenge at best, and this is where Supervised Machine Learning comes into play. }Table~\ref{tab:How2DoRegressionReduced} shows a vanilla implementation where the \textit{features} are selected as time and the $x_1$-states and the \textit{labels or targets} are taken as the inputs $u$ and the $x_2$-states. In the control implementations simulated here, the features are taken as the Cartesian hip velocities only (the positions are discarded) and the labels are taken as the outputs listed in Table~\ref{table:ik_mapping}.

\begin{table*}
	\centering
    	\caption[]{Conceptual arrangement of the optimization data from which appropriate functions for building a stabilizing controller can be determined by ``regression''. Here $\xi^i$, $1 \le i \le M$ is a discretization of $Z_0$, $t_j$, $1 \le j \le N$ is a discretization of $[0, T_p]$, \add{$\varphi_{\xi}(t)$ is the solution of the full model with the initial condition $\xi$ at $t=0$, $\pi_1$ is the projection onto the $x_1$ coordinate, $\pi_2$ is the projection onto the $x_2$ coordinates, and $u_{\xi}(t)$ is the input giving rise to the solution $\varphi_{\xi}(t)$}. Standard toolboxes in MATLAB are used to build the functions and even to test for their existence via the quality of fit tools.}
		\begin{tabular}{c|c||c|c}
			\multicolumn{2}{c||}{\bf Features } &   \multicolumn{2}{c}{ \bf Labels or Targets}  \\
			\hline
			$t_j$& $x_{1}^{j, i}=\pi_1 \circ \varphi_{\xi^i}(t_j) $ & $\nu^{j, i}=\pi_2 \circ \varphi_{\xi^i}(t_j)$  & $\mu^{j, i}=u_{\xi^i}(t_j)$\\
			\hline
            & & & \\
			$t_0=0$ & $x_{1}^{0,1} $  & $\nu^{0,1}$ &  $\mu^{0,1}$\\
            & & & \\
			$t_0=0$ & $x_{1}^{0,2} $  & $\nu^{0,2}$ &  $\mu^{0,2}$ \\
			$\vdots$ & $\vdots$& $\vdots$ & $\vdots$ \\
			$t_0=0$ & $x_{1}^{0,M} $  & $\nu^{0,M}$  &  $\mu^{0,M}$\\
            & & & \\
			$t_1$ & $x_{1}^{1,1} $  & $\nu^{1,1}$  &  $\mu^{1,1}$\\
			$\vdots$ & $\vdots$& $\vdots$ & $\vdots$ \\
			$t_1$ & $x_{1}^{1,M} $  & $\nu^{1,M}$ &  $\mu^{1,M}$ \\
			$\vdots$ & $\vdots$& $\vdots$  &$\vdots$  \\
			$t_N=T_p$ & $x_{1}^{N,1} $  & $\nu^{N,1}$  &  $\mu^{N,1}$\\
			$\vdots$ & $\vdots$& $\vdots$ & $\vdots$  \\
			$t_N=T_p$ & $x_{1}^{N,M} $ & $\nu^{N,M}$ &   $\mu^{N,M}$\\
		\end{tabular}
	\label{tab:How2DoRegressionReduced}
\end{table*}

\begin{table*}
\centering
  \caption{Table showing a mapping between higher-level control objectives and their realization at a lower-level.}
  \label{table:ik_mapping}
  \begin{tabular}{ |p{5cm}|p{8cm}| }
    \hline
    Outputs & Actuated Joints \\
    \hline
    \hline
    Torso Pitch, Roll, and Yaw orientation & Stance Hip Pitch, Roll, and Yaw Joints\\
    \hline
    Stance Knee Joint & Stance Knee Joint \\
    \hline
    Swing Knee Joint & Swing Knee Joint \\
    \hline
    Stance Pitch Ankle Joint & Stance Pitch Ankle Joint \\
    \hline
    Stance Henke Ankle Joint & Stance Henke Ankle Joint \\
    \hline
    Swing Hip Pitch Joint & Swing Hip Pitch Joint \\
    \hline
    Swing Hip Roll Joint & Swing Hip Roll Joint \\
    \hline
    Swing Foot Pitch, Roll, and Yaw orientation & Swing Pitch and Henke Ankle Joints, and Swing Hip Yaw Joint \\
    \hline
  \end{tabular}
\end{table*}

\begin{figure}
	\centering
	\includegraphics[width=1\columnwidth]{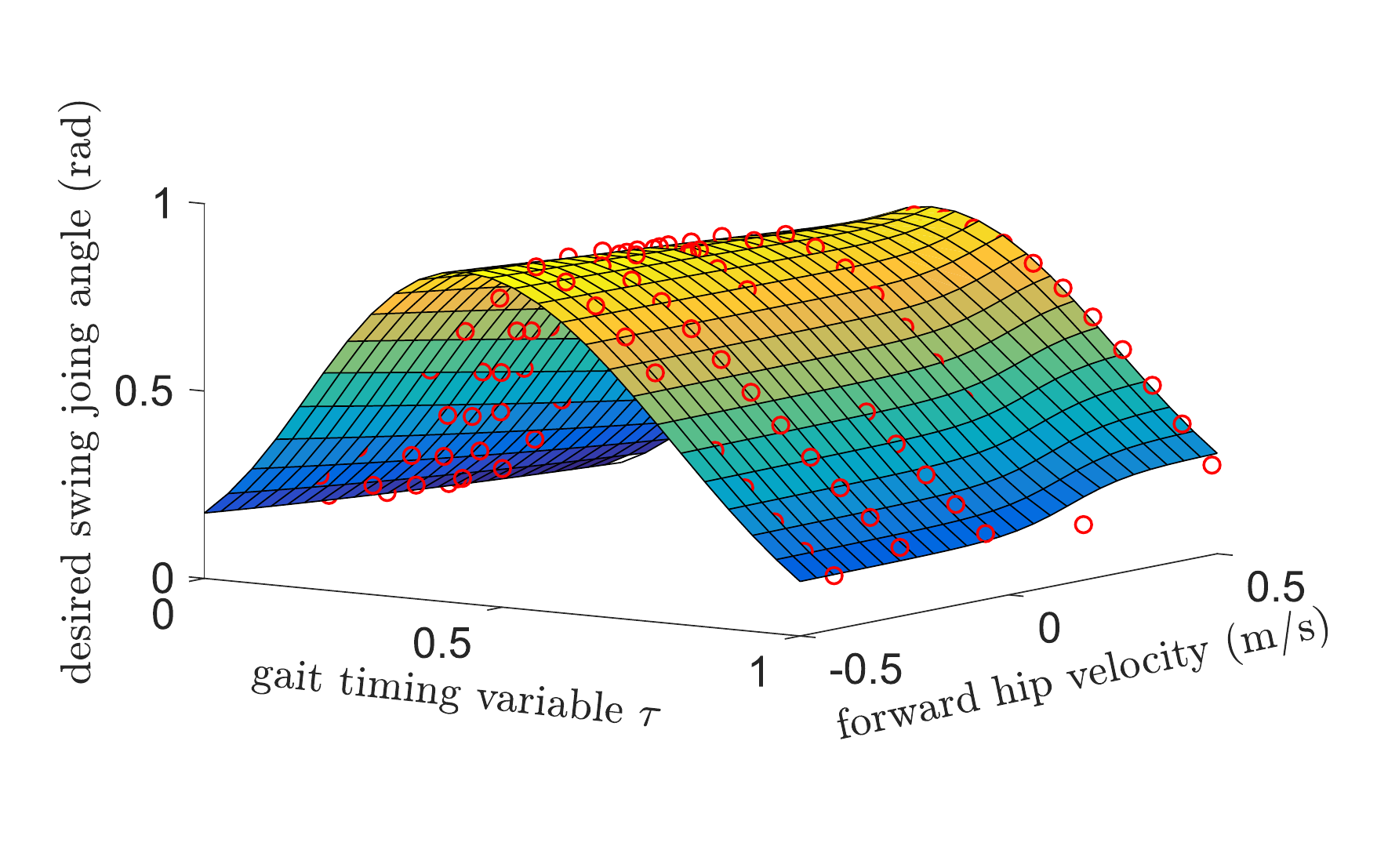}
	\caption{Surface showing the desired swing knee angle as a function of the gait timing variable $\tau$ and the forward hip velocity for stabilizing the step-in-place gait. The red circles are the actual data obtained from optimization while the surface is a regression done using the Neural Network Toolbox in MATLAB. Note that some of the circles are under the surface and are therefore not visible.}
    \label{fig:swing_knee_angle}
\end{figure}

At this point we have all the functions we need to implement a control policy that locally exponentially stabilizes the selected periodic walking gait. \figref{fig:swing_knee_angle} shows one component of the function $\nu$ arising from the Supervised Machine Learning.

\subsection{Step 5: Feedback Control to Render $Z$ Attractive.}
\begin{figure}
	\centering
	\includegraphics[width=0.8\columnwidth]{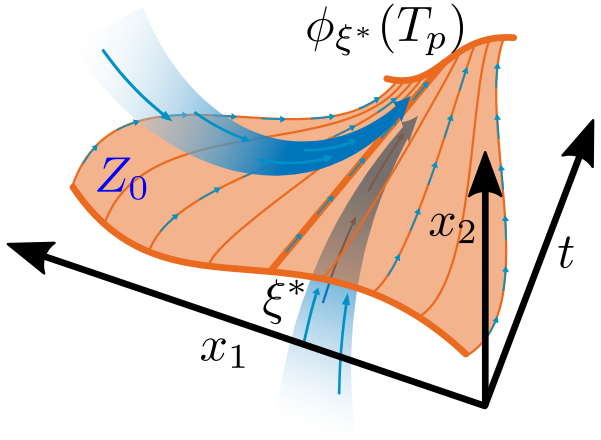}
   	\caption{The low-dimensional realization (invariant surface and vector field) being rendered attractive in the full order model.}
    \label{fig:MLstep3}
\end{figure}

The previous steps created a low-dimensional dynamical model for which the desired periodic orbit with initial condition $\xi^* \in Z_0$ is locally stable and attractive in $Z$.  The next step is to stabilize the orbit in the full model, \add{as illustrated in Figure~\ref{fig:MLstep3},} and for this we need to be more specific about the $x_2$-portion of the model.  In the case of the exoskeleton, the strongly actuated part of the model is fully actuated and is therefore feedback linearizable  \cite{SP94}. In particular, the $x_2$-part of the model \eqref{eq:Decomposed Model} can be expressed as 
\begin{align}
&x_2=\left[ \begin{array}{c} x_{2a} \\ x_{2b}  \end{array} \right], \nonumber 
\end{align}
and
\begin{align}
\label{eq:Decomposed ModelSpong}
&f_2= \left[ \begin{array}{c} x_{2b} \\ \alpha(x_1,x_2) + \beta_1(x_1,x_2) u_1 + \beta_2(x_1,x_2) u_2 \end{array} \right],
\end{align}
where $\beta_2$ is square and invertible.  

In \cite{IJRR_DaGrizzle2017}, it is shown that for all  $\frac{n_2}{2} \times \frac{n_2}{2} $ positive definite matrices $K_p$ and $K_d$, $\varphi_{\xi^*}:[0, T_p) \to \real^n$ is a locally uniformly exponentially stable solution of the closed-loop system
\begin{equation}
\label{eq:ModelWithFeedback}
\begin{aligned}
&\dot{x}_1= f_1(x_1,x_2,u_1)\\
&\dot{x}_2= f_2(x_1,x_2,u_1,u_2) \\
&u_1=v_1 \\
&u_2= \left[ \beta_2(x_1,x_2) \right]^{-1} \big(-\alpha(x_1,x_2) -\beta_1(x_1,x_2)u_1+ v_2 \big) \\
&\begin{bmatrix}v_1 \\ v_2 \end{bmatrix} =  \mu(t,x_1) - [K_p~~K_d]\big(x_2 - \nu(t, x_1) \big).
\end{aligned}
\end{equation}

\subsection{Step 6 (Optional): Enriching the Control Policy to Handle Multiple Gaits }

The final step performed with the ATALANTE model is to repeat the above processes for a grid of periodic gaits corresponding to a range of walking speeds captured in $Z_0$. The trajectory designs for these new gaits are catenated to Table~\ref{tab:How2DoRegressionReduced}, with the only change being that the feature set is augmented to include the designed average (Cartesian) velocity of each gait. Supervised Machine Learning, if successful, then produces a control policy that allows a desired walking speed to be selected.

\subsection{Remark on Trajectory Design via Optimization}
\begin{figure}
	\centering
    \includegraphics[width=0.9\columnwidth]{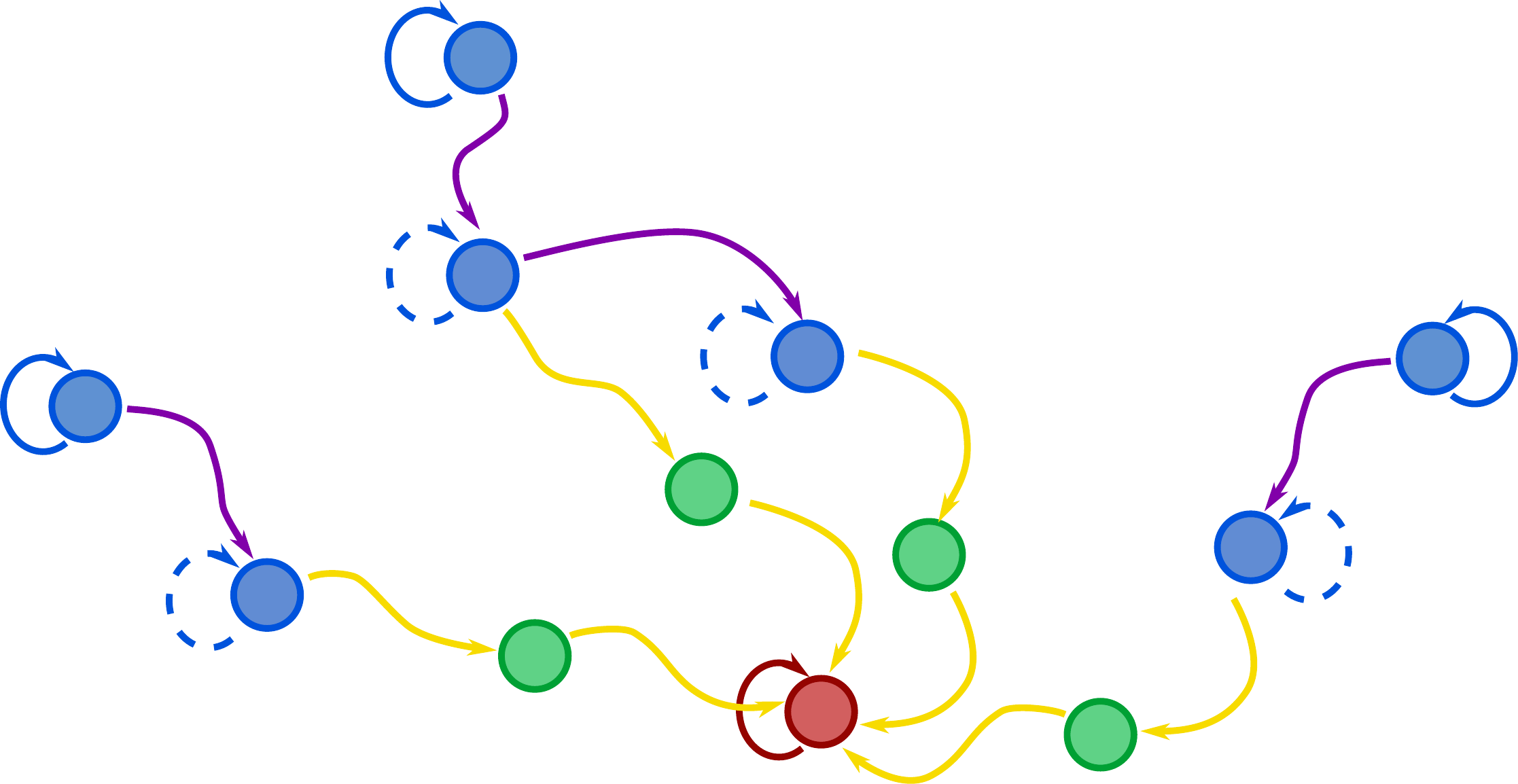}
	\caption{Graphical representation of gait optimization. Each arc represents one step of the exoskeleton. Circles with self-loops denote periodic orbits, while those without are transient states. The red circle denotes the goal (periodic) gait. The optimizations are done over three steps, with the end of the first step required to terminate in $Z_0$, which is parameterized by periodic gaits, and the end of the third step must be the goal state. Since the first step initiates and ends in $Z_0$, the three-step optimization can continue from the end of a first step and other self-loop-states in the same fashion as described above. The (first-step) trajectories corresponding to the time interval $[0, T_p]$---denoted by the purple arcs---are saved and used as training data for the Supervised Machine Learning. The other data is discarded.}
    \label{fig:transition_opt_instruction}
\end{figure}

We provide a few details on how to actually generate the trajectories in \eqref{eq:PeriodicBehavior} that form the surface in \figref{fig:MLstep1}. We use optimization to determine solutions of the model; see ``\nameref{sidbar:direct_collocation}'' for more details on the optimization method itself. In particular, a cost function of the form
\begin{equation}
	\begin{aligned}
		J(\xi)=& \min_{u_\xi} \int_{0}^{3T_p} L( \varphi_\xi(\tau), u_\xi(\tau)) d \tau \\
		\text{s.t.}~{\varphi}_\xi(t) = & \xi + \int_{0}^{t}f(\varphi_\xi(\tau),u_\xi(\tau)) d \tau \\
		0 \ge & c( \varphi_\xi(t), u_\xi(t)),~0 \le t \le 3 T_p\\
		\varphi_\xi(T_p)\in & Z_0,~~||\varphi_\xi(T_p) - \xi^*|| \le c ||\xi - \xi^*|| \\
		\varphi_\xi(3 T_p) = & \xi^*
	\end{aligned}
	\label{eq:CostFunction}
\end{equation}
with constraints is posed over a time horizon of three steps such that (i) energy per step taken is penalized, (ii) solutions satisfy the full-order model, (iii) key constraints on ground reaction forces and actuator limits are respected (captured in the function $c \le 0$  ), (iv) solution returns to $Z_0$ \add{in one step} so that trajectories can be continued indefinitely, while approaching the desired periodic orbit, and (v) terminates at the nominal periodic orbit in three steps (this could be replaced by a terminal penalty in the cost function). \add{The choice of three (robot) steps in the optimization is based on the capture-point analysis in \cite{ZaHaRu15} and theoretical work in \cite{IJRR_DaGrizzle2017}}. \add{$L( \varphi_\xi(\tau), u_\xi(\tau))$ defines the running cost while $f(\varphi_\xi(\tau),u_\xi(\tau))$ is the dynamics of the system.} \figref{fig:transition_opt_instruction} shows a conceptual representation of the gait design process.

%% file: section_numerical_validation.tex
\section{Numerical Illustration}

This section presents simulation results of the G-HZD controller developed in the previous section. The simulation experiments were done with Gazebo using the full dynamical model of the exoskeleton with a human in it. The ground contact model in Gazebo allows the possibility for the feet to roll and slip. \addb{In these tests, the machine learning approach based on G-HZD was used to stabilize walking motions with a longitudinal speed range of -0.3 m/s to 0.3 m/s.}


\subsection{Velocity Tracking}
\begin{figure}
	\centering
    \includegraphics[width=0.8\columnwidth]{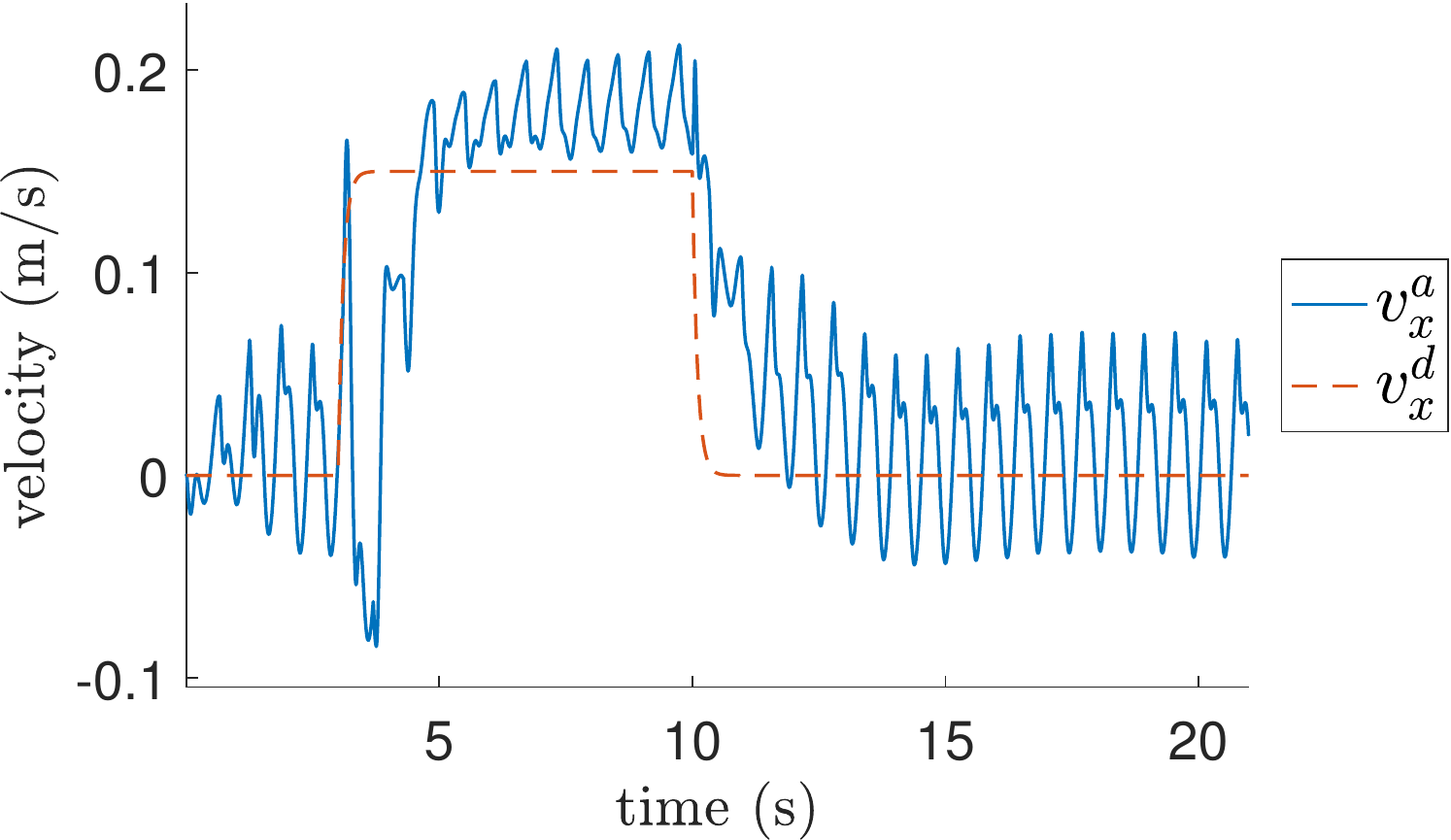}
	\caption{Shows tracking of a desired velocity profile. $v^a_x$ is the actual forward velocity of the pelvis during walking. $v^d_x$ is the commanded average forward velocity.}
    \label{fig:vel_tracking}
\end{figure}

\begin{figure}
	\centering
    \includegraphics[width=0.8\columnwidth]{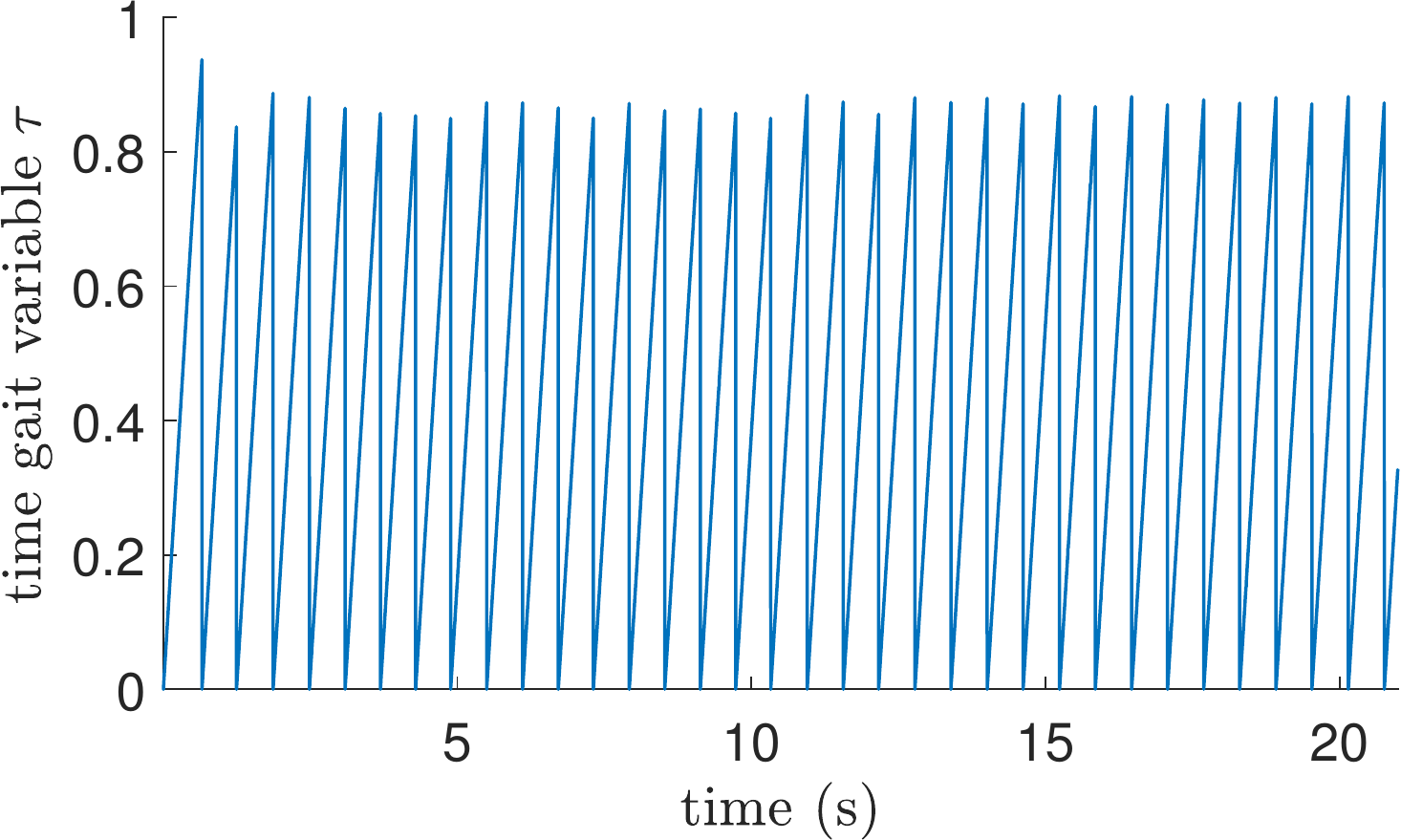}
	\caption{Shows the gait timing variable $\tau$ changing as a function of time while the exoskeleton was following a varying desired speed.}
    \label{fig:vel_track_tau}
\end{figure}

\begin{figure}
  \centering
  \subcaptionbox{Swing Sagittal Hip Phase Portrait}
  {
    \includegraphics[width=0.45\columnwidth]{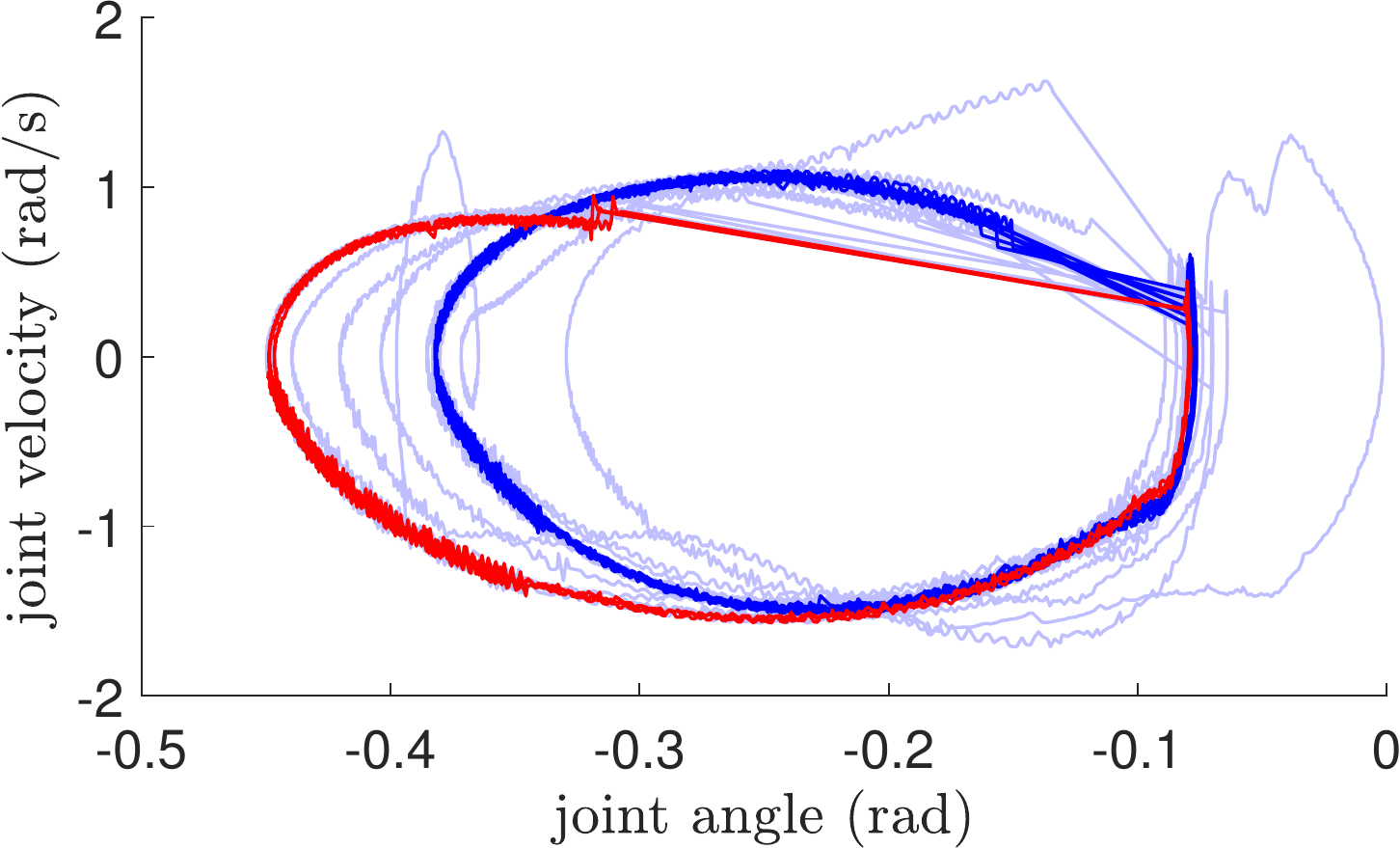}
  }
  \subcaptionbox{Swing Knee Phase Portrait}
  {
    \includegraphics[width=0.45\columnwidth]{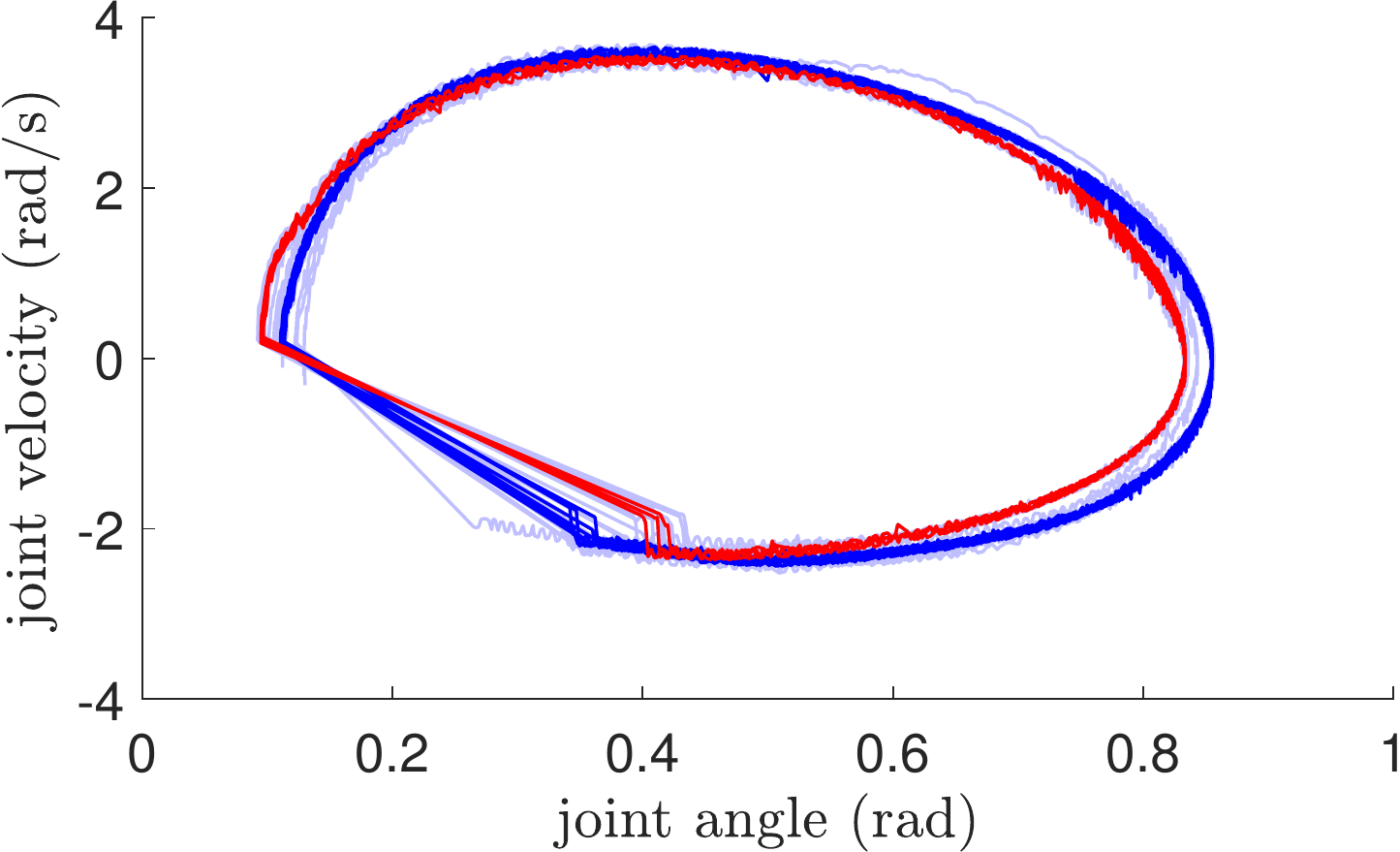}
  }
  \caption{The phase portrait of both the swing and stance sagittal hip joints for when the system was tracking a desired velocity profile. Shows $q$ vs $\dot{q}$ for the two joints. The solid blue line shows the states at the last quarter of the simulation. While the solid red line shows the part of the simulation where the exoskeleton reached a steady state trajectory while walking forward. It can be seen that both cases, the exoskeleton converged to a periodic cycle.}
  \label{fig:vel_track_phase_portrait}
\end{figure}

\begin{figure*}
	\centering
    \includegraphics[width=0.8\linewidth]{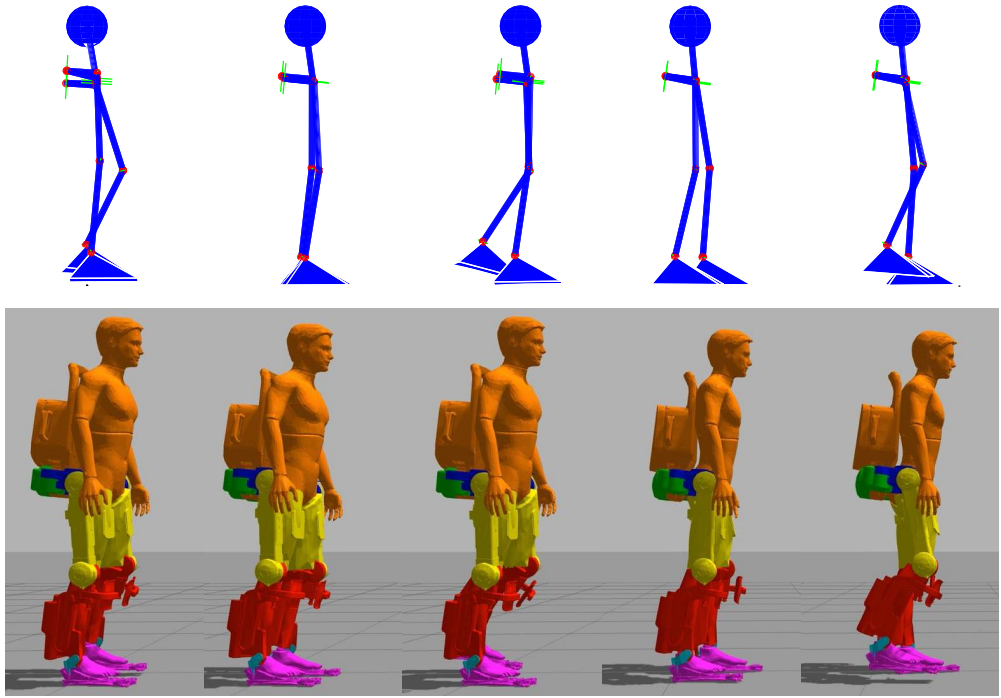}
	\caption{Outtakes from animations of the exoskeleton tracking a desired velocity profile. Below are images from the Gazebo simulation, while above are stick figures with the same motion added for clarity. The first two photos show the step-in-place gait. The third image shows how the system looks like when it is starting to transition. The last two show the exoskeleton walking forward in a periodic gait.}
    \label{fig:vel_track_stick}
\end{figure*}

\figref{fig:vel_tracking} shows a simulation where the exoskeleton starts with stepping in-place at $t=0$~s, then at $t=3$~s, the exoskeleton is commanded to walk at a speed of $0.15$~m/s, until at $t=10$~s, the exoskeleton is commanded to return to a stepping in place gait. It can be seen that there is a small steady-state error when forward walking is commanded; this can be attributed to a combination of joint tracking errors from the low-level PD-controllers and the compliant (non-rigid) ground being different from the model used for control design. \figref{fig:vel_track_tau} shows the gait timing variable exhibiting a typical triangle-wave pattern, with leg swapping happening before $\tau = 1$. It can be seen that regardless of whether the system was in a periodic gait or was transitioning between speeds, $\tau$ consistently terminates at around 0.9. This is due to having trained the system on various transition gaits. The phase portrait given in \figref{fig:vel_track_phase_portrait} shows the periodic nature of stepping in-place, followed by a transient to another periodic condition. Convergence to periodic motion is clear during periods when the speed command is constant. Stick figures showing stepping-in-place, transitioning, and walking forward can be seen in \figref{fig:vel_track_stick}. An animation of the simulation result can be seen in \cite{simVideo}.

\subsection{Preliminary Robustness Analysis}

\addb{It is easy to imagine scenarios that could lead to an exoskeleton losing stability and ``tripping up'', including unexpected contact with objects in the environment, walking over uneven terrain, external force perturbations being applied on the system, or spasticity
(i.e., involuntary muscle resistance to patient's leg motion).
In previous work, we evaluated in simulation a PHZD-based controller of the exoskeleton in the face of unexpected slopes and unplanned upper-body motion \cite{Agrawal2017First}.
Here, we present a preliminary robustness analysis of the machine learning controller under various external and internal perturbations.}

\begin{figure*}
  \centering
  \subcaptionbox{Supervised Machine Learning}
  {
    \includegraphics[width=0.4\linewidth]{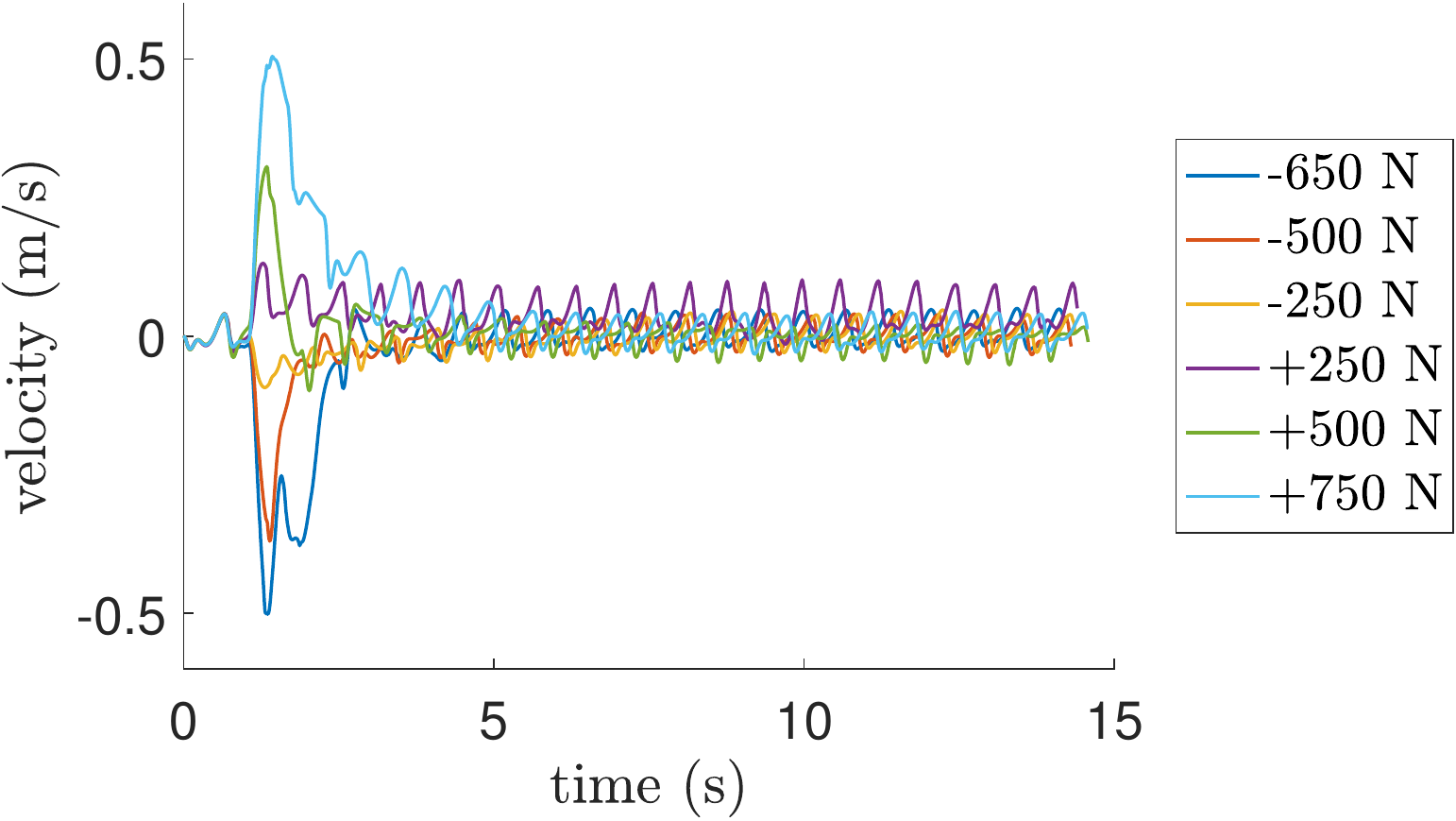}
  }
  \subcaptionbox{Fixed Gait}
  {
    \includegraphics[width=0.4\linewidth]{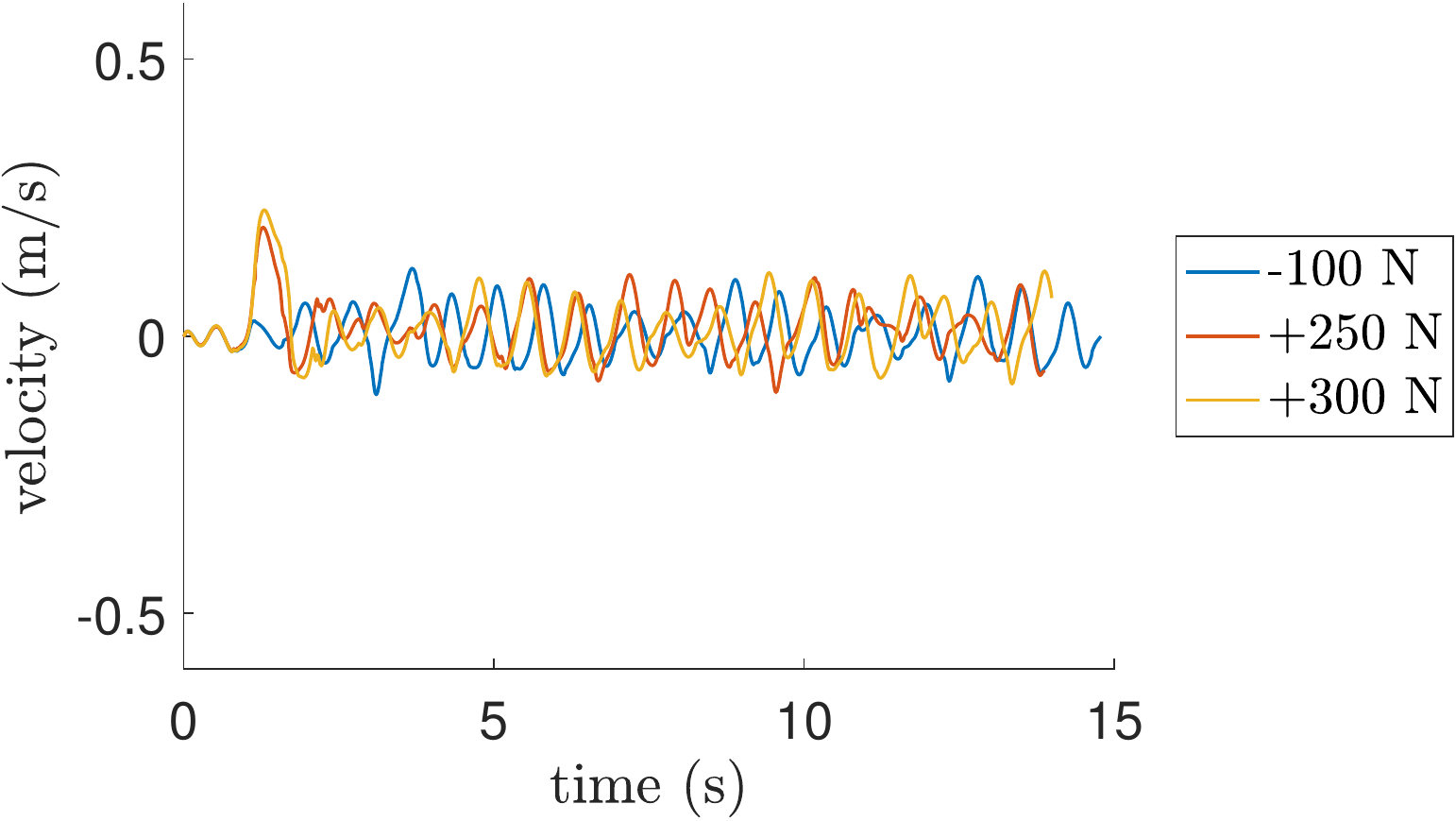}
  }
  \caption{Change in hip velocity due to a perturbation. The control algorithm based on Supervised Machine Learning is able to recover from larger perturbations. Both controllers failed under perturbations larger than those shown.}
  \label{fig:force_vel}
\end{figure*}

\begin{figure*}
  \centering
  \subcaptionbox{Supervised Machine Learning}
  {
    \includegraphics[width=0.4\linewidth]{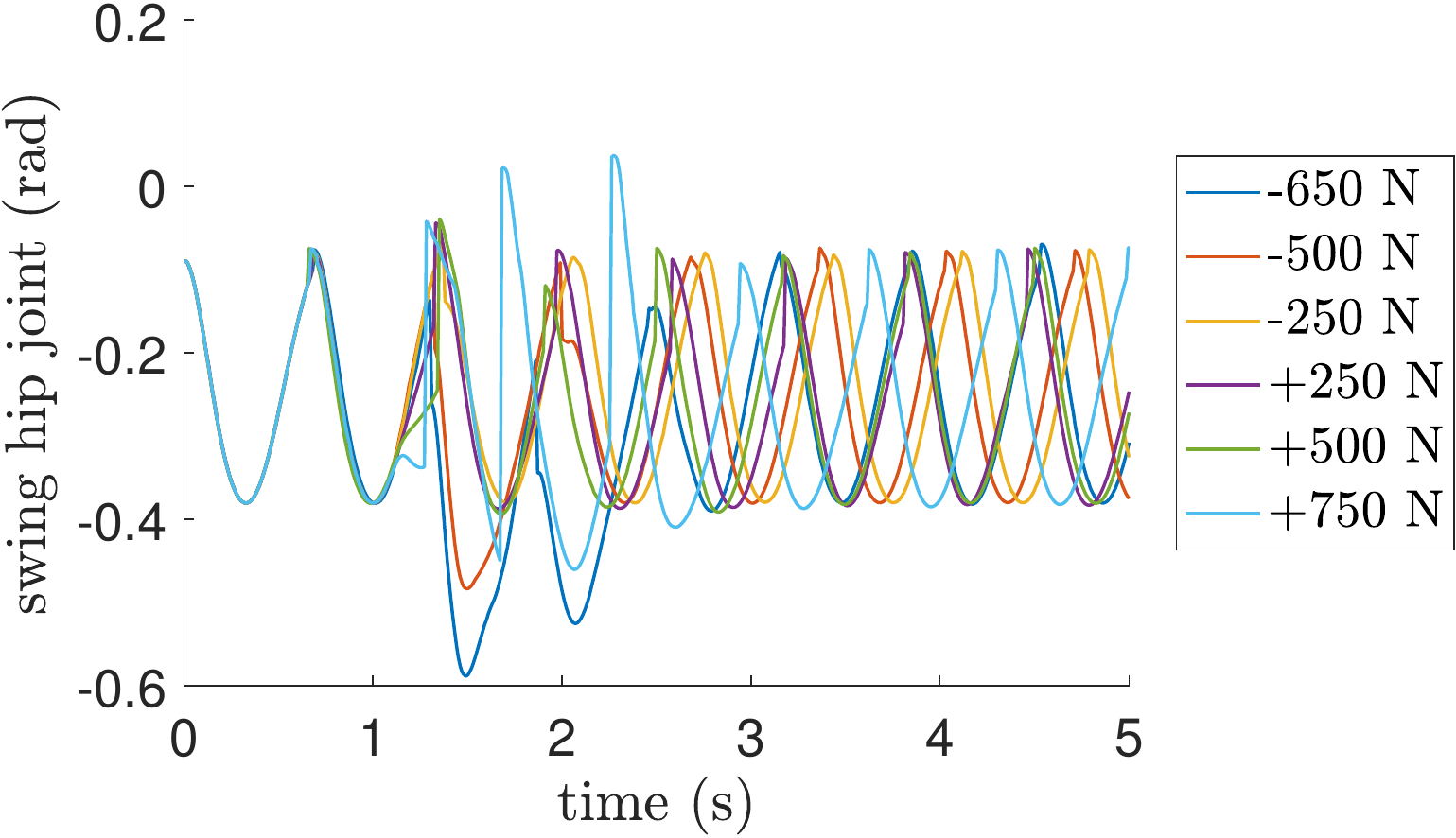}
  }
  \subcaptionbox{Fixed Gait}
  {
    \includegraphics[width=0.4\linewidth]{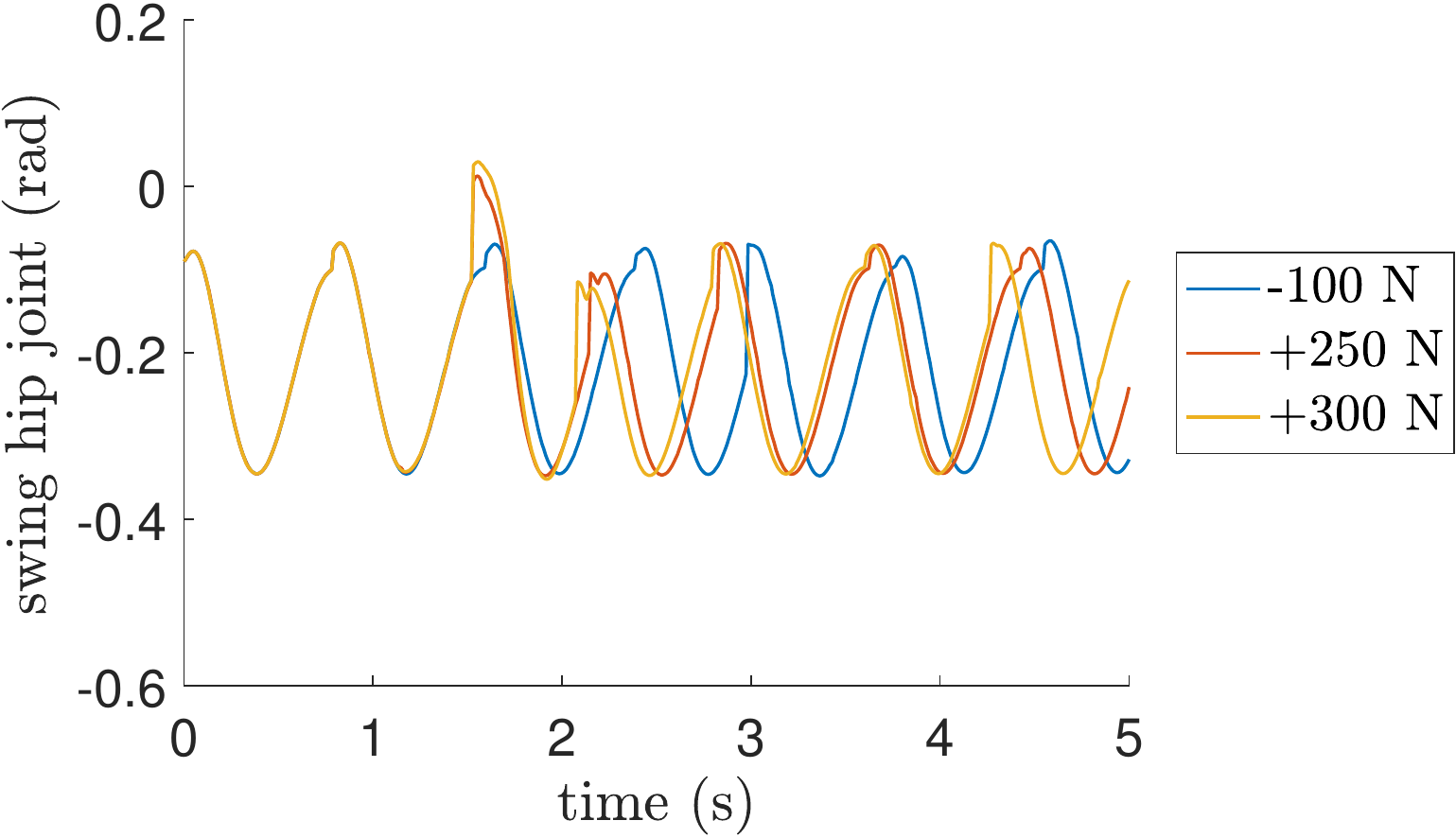}
  }
  \caption{The swing hip joint angle as a function of time while responding to a perturbation. When experiencing large perturbations, the controller based on Supervised Machine Learning makes more use of its swing foot in order to return to a stepping-in-place gait.}
  \label{fig:force_swinghip}
\end{figure*}

\newsec{\addb{Velocity Perturbations.}} \addb{Analysis in \cite{griffin2016nonholonomic} shows that a bipedal robot's ability to reject velocity perturbations when using HZD-based control strategies correlates with its ability to reject a set of other perturbations, such as variations in terrain height. Based on this observation, it is posited that velocity perturbation rejection provides a reasonable preliminary test of the robustness of the proposed controller framework.}
In this test, we induce velocity perturbations by applying impulsive external forces of various magnitudes and directions to the exoskeleton while it is walking. We compare the response of two different closed-loop configurations: one with a controller using the machine learning and another with a fixed single periodic gait. The external force is applied at the hip of the exoskeleton during the second step for $0.1$ seconds in either forward or backward directions.
When the exoskeleton is commanded to step in place, the results of Gazebo simulation are shown in \figref{fig:force_vel} and \figref{fig:force_swinghip}. In particular, using the machine learning controller, the exoskeleton is able to recover from up to 750 N force in the forward direction and 650 N force in the backwards direction, whereas, when using a controller for fixed periodic gait, the system lost stability with any forward force larger than 300 N or a backward force larger than 100 N. \figref{fig:force_vel} shows the changes in the forward hip velocity of the exoskeleton for both machine learning and fixed-gait methods. It is seen that the Supervised Machine Learning controller is capable of recovering for relatively large velocity disturbances arising from external forces. In addition, \figref{fig:force_swinghip} shows how the controllers respond to the external disturbances by changing the desired trajectories of the exoskeleton joints. Specifically, the machine learning controller, when challenged with a large perturbation, extends the swing leg outward by modifying the desired trajectory of the sagittal swing hip joint. This behavior occurs naturally from training the control surface on optimized walking gaits for various speeds and transitions among them. The fixed-gait controller only uses the large feet of the exoskeleton (regulating the ZMP) to reject the perturbation instead of adjusting the step length. An animation of the velocity perturbation simulations in Gazebo can be seen in \cite{simVideo}.

\begin{figure}
  \centering
  \includegraphics[width=0.95\columnwidth]{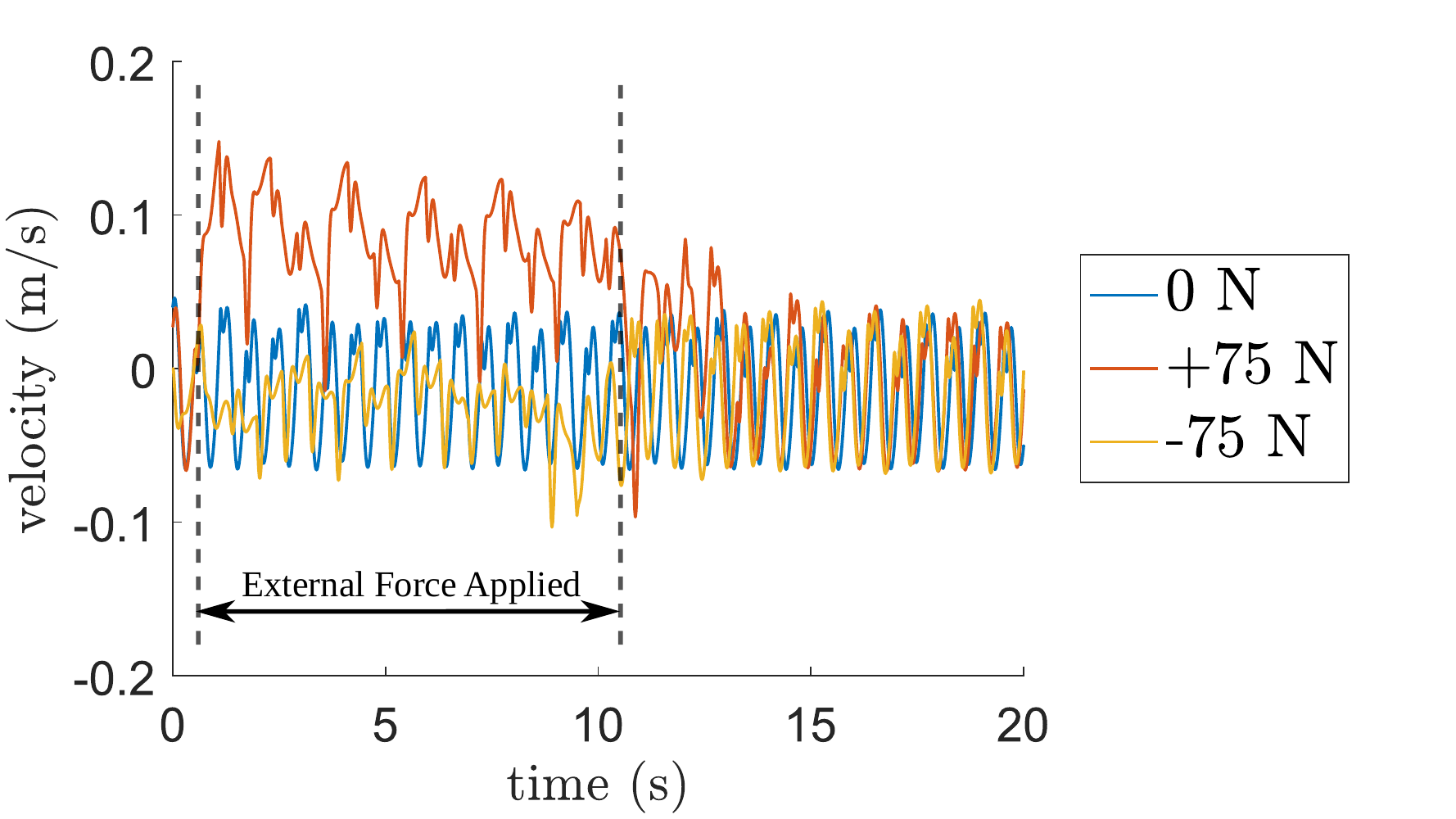}
  \caption{Instantaneous velocity of the hip when system is perturbed for a duration for 10 seconds. The forward push causes the system to drift slightly more while the force is being applied. When force is removed, the velocity converges back to the nominal velocity.}
  \label{figure:vel_const_force}
\end{figure}

\newsec{\addb{Constant Force Perturbation.}} \addb{While studying velocity perturbation under impulses captures short duration events, such as someone or something bumping into the exoskeleton, it is also interesting to investigate the effects of persistent external forces applied to the system. Figure~\ref{figure:vel_const_force} shows the average speed when a constant force is applied to the torso, 0.5 m above the pelvis, under the controller based on Supervised Machine Learning. The exoskeleton is able to withstand a constant force of $\pm$75 N for a duration of 10 seconds. Small drifts in velocity are noticed with the forces applied. The velocities converge to the nominal velocity after the force is removed. Larger forces result in loss of stability.}

\newsec{\addb{Unplanned forces in the user's legs.}} \addb{A simplified model of spasticity is represented by torques at a user's knee joints. While spasticity occurring simultaneously in both legs was investigated, only spasticity in the right leg is reported here. Both constant and sinusoidal torques are applied, and these are added to the joint-side torque provided by the knee motor and gearing. The mean-absolute error (MAE) in joint tracking is computed for the right knee. It is observed that the MAE increases only slightly, as shown in Table~\ref{table:spasm_results}. In addition, no differences in the gaits is visually observed.}

\begin{table}[]
	\renewcommand{\arraystretch}{1.3}
	\caption{The mean-absolute error (MAE) for joint angle tracking when applying external torques on the right knee joint representing a simplified model of spasticity.}
  	\label{table:spasm_results}
	\begin{centering}
		\begin{tabular}{c|c}
        	\textbf{Test} & \parbox{2cm}{\centering \textbf{MAE (rad)} \\ \textbf{(at steady-state)}} \\
			\hline 
			No perturbations & 0.0130\\
			Constant +100 Nm torque & 0.0166\\
			Constant -100 Nm torque  & 0.0180\\
			Sinusoidal 100 Nm torque at 1 Hz frequency & 0.0147\\
			Sinusoidal 100 Nm torque at 10 Hz frequency & 0.0154\\
		\end{tabular}
	\par
    \end{centering}
\end{table}

%% file: section_phzd_experiments.tex
\section{Preliminary Experimental Results using PHZD}

Experimental implementation of the biped-inspired control laws has begun, with very promising results as alluded to in the opening lines of this article. Because the PHZD control laws have been extensively evaluated on several bipedal robot platforms, they have been employed in the initial testing. The results described below were first reported in \cite{Gurriet2018Towards}, where a video of patients using the exoskeleton can be found. 

\begin{figure*}
	\begin{centering}
		\includegraphics[width=0.8\linewidth]{\useSD{figure_20a.png}}\\
		\includegraphics[width=0.8\linewidth]{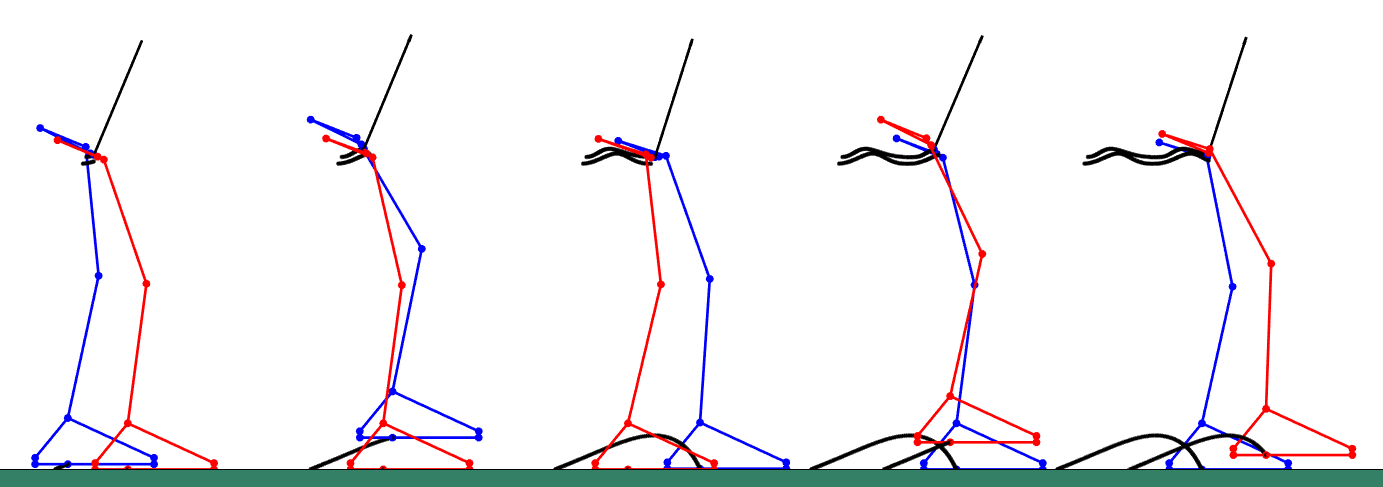}
		\par
    \end{centering}
	\caption{\add{Tiled snapshot images of the nominal gait from the optimization for the exoskeleton with a mannequin inside in experiment and simulation. The black lines in the simulation images shows the evolution of the pelvis and the swing sole positions.}}
    \label{fig:SimuTiles}
\end{figure*}

\begin{figure}
\begin{centering}
\includegraphics[width=1.0\columnwidth]{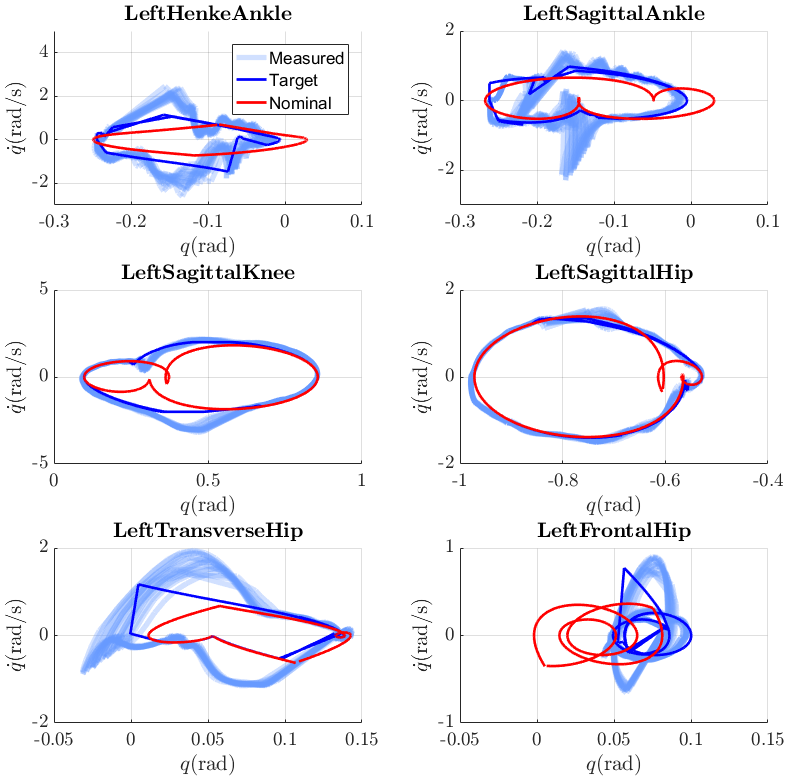}
\par\end{centering}
\caption{Phase portraits for the dummy during 20 s of unassisted walking \cite{Gurriet2018Towards}. \add{Red is the nominal orbit from optimization while blue is the ``tuned'' orbit to accommodate for mechanical imperfections
and compliances of the robot.} \label{fig:PhaseD}}
\end{figure}

In the first evaluation of the controllers, a mannequin or dummy was placed in the exoskeleton as shown in \figref{fig:SimuTiles}.
As we can see in \figref{fig:PhaseD}, the nominal and target trajectories \add{(in red and blue respectively)} are marginally different after the tuning and high-level filtering of the nominal trajectories. 
The target gait is followed with relatively good accuracy resulting in stable dynamic walking of the hardware.

\subsection{Experimental results with human subjects}
As a result of the successful results obtained with the mannequin, experiments are carried out with paraplegic patients.
\add{All patients are complete paraplegic, unable to stand by themselves, unable to walk, with lesion ranged from T12 to T6.}
Some characteristics of these patients are summarized in Table~\ref{tab:PatientData}.

\begin{table}
	\renewcommand{\arraystretch}{1.3}
	\caption{Patient data. \add{Selection criteria included being unable to stand by oneself, able to remain seated on a chair without further assistance, no ambulatory ability before using the exoskeleton, and low-to-moderate spasticity allowed.}}
    \label{tab:PatientData}
	\begin{centering}
		\begin{tabular}{c|c|c|c|c}
			\textbf{Patient} & \parbox{1cm}{\centering \textbf{Height} \\ \textbf{(m)}} & \parbox{1cm}{\centering \textbf{Weight} \\ \textbf{(kg)}} &  \parbox{1.5cm}{\centering \textbf{Distance} \\ \textbf{traveled (m)}} & \parbox{1cm}{\centering \textbf{Speed} \\ \textbf{(m/s)}}
			\tabularnewline
			\hline 
			A & 1.80 & 68 & 8.9 & 0.11 \\
			B & 1.69 & 80 & 10.56 & 0.15\\
			C & 1.80 & 75 & 9.5 & 0.13\\
		\end{tabular}
	\par
    \end{centering}
\end{table}

Experiments were conducted in a certified medical center and approved by the ANSM (French regulatory administration for health products). To prevent injury from a fall, one person is placed at each side of a patient. In case of loss of balance, the two assistants catch the exoskeleton by handles on its sides. Furthermore, a safety cable is attached to the exoskeleton and an overhead rail (or gantry). This is a secondary means to secure a patient and prevent a fall. To be clear, assistance is provided only in case of loss of balance. During walking, the exoskeleton and its user are self-stabilized and no outside assistance was given.

\begin{figure*}
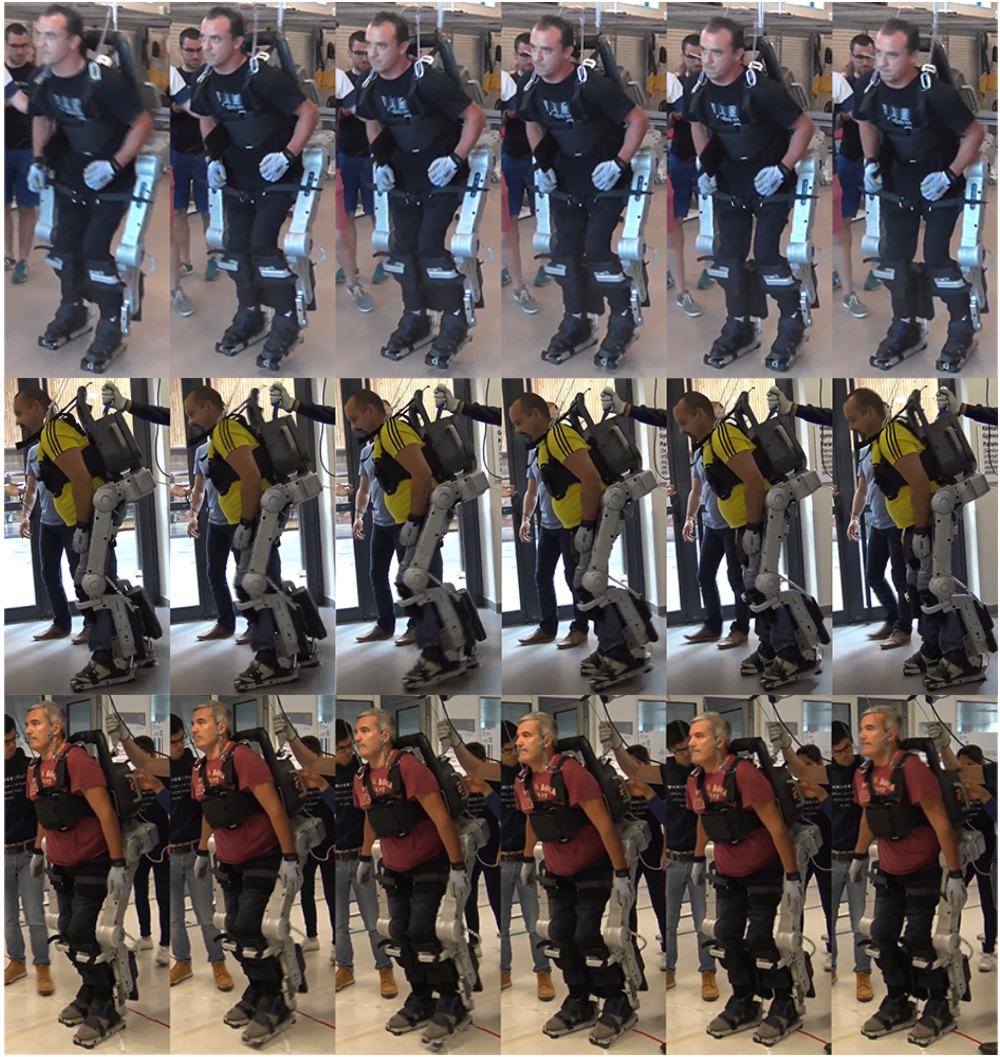

  \centering
 \includegraphics[width=0.8\linewidth]{\useSD{figure_22a.png}}\\
 \includegraphics[width=0.8\linewidth]{\useSD{figure_22b.png}}\\
 \includegraphics[width=0.8\linewidth]{\useSD{figure_22c.png}}
  \caption{From top to bottom, walking tiles of patients A, B and C.}
  \label{fig:TileXPs}
\end{figure*}

As can be seen in \figref{fig:TileXPs}, which shows tiles from the video of  \cite{Gurriet2018Towards}, crutch-less dynamically stable exoskeleton walking of paraplegic patients is achieved as a result of the methodology developed for bipedal robots. All patients managed to walk unassisted for the entire length of the room after a few trials during which a best gait was chosen and then tuned; Table \ref{tab:PatientData} includes the speed of walking and distance traveled.

\begin{figure}
  \begin{centering}
  	\includegraphics[width=1.0\columnwidth]{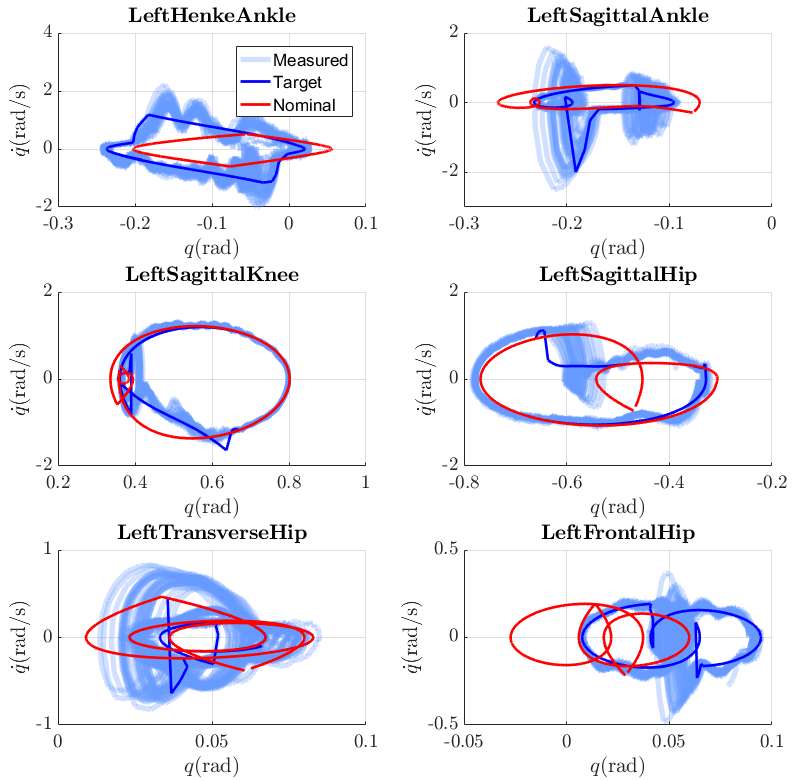}
    \par
  \end{centering}
  \caption{Phase portraits for patient A during 60 s of unassisted walking.}
  \label{fig:PhaseA}
\end{figure}

\begin{figure}  
\hspace{-5mm}
  	\includegraphics[width=1.1\columnwidth]{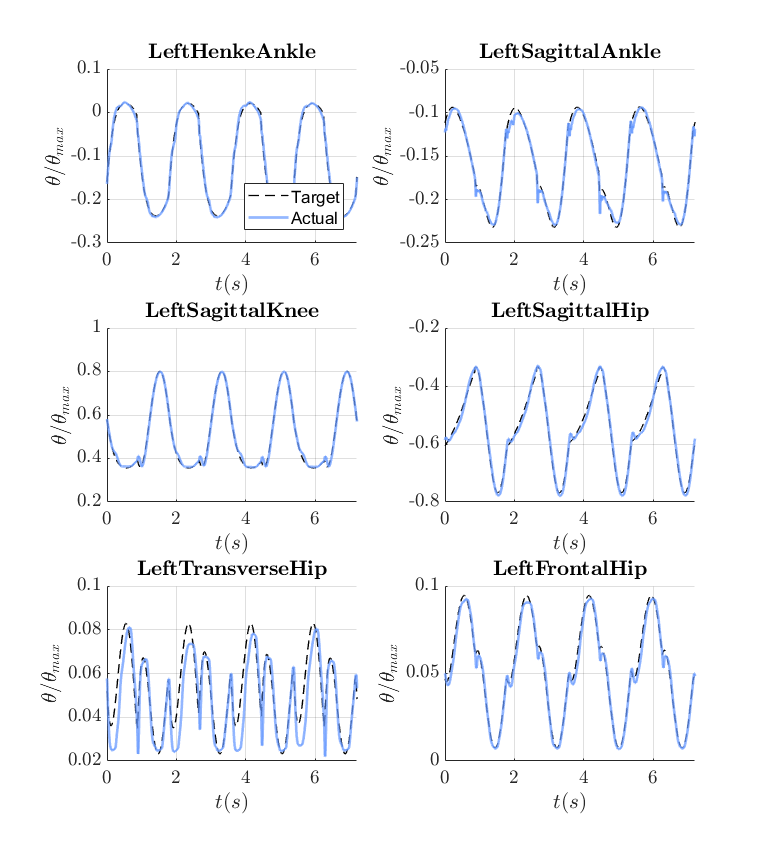}
    \par
  \caption{\add{Selection of normalized tracking performance of the local controllers at the joint for patient A.}}
  \label{fig:TrackingA}
\end{figure}

\begin{figure}
  \centering
  \includegraphics[width=0.98\columnwidth]{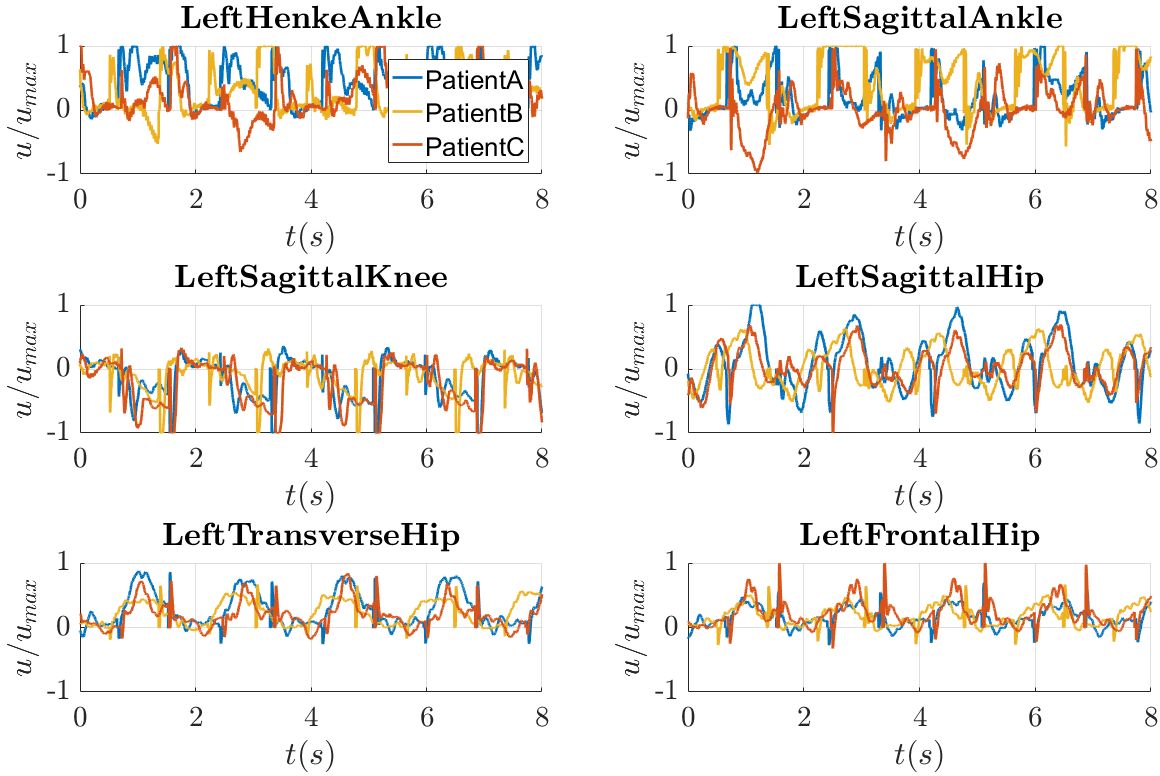}
  \caption{\add{Normalized experimental motor torques for three different patients \cite{Gurriet2018Towards}. Other that for patient B, the motor torques rarely saturate. Hence there is reserve torque for responding to disturbances.}}
  \label{fig:Torques}
\end{figure}

\begin{figure}
  \centering
  \includegraphics[width=0.98\columnwidth]{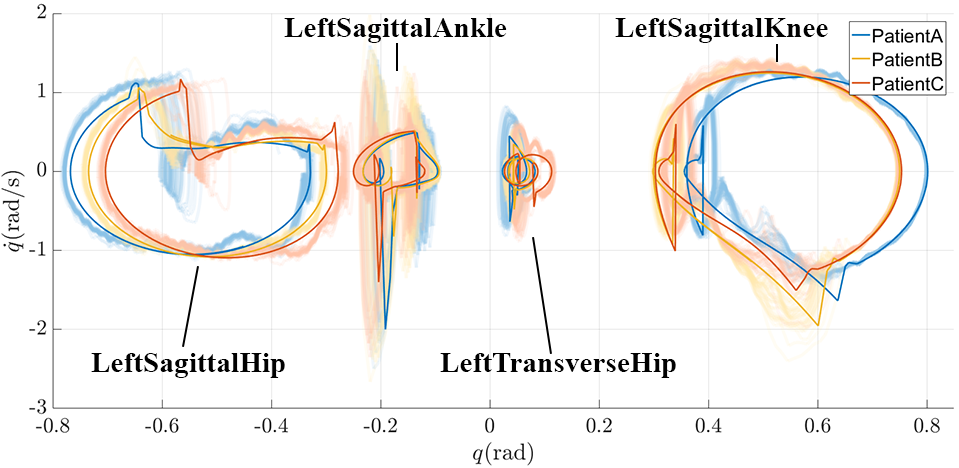}
  \caption{Selection of phase portraits. The solid lines are the target trajectories. The shaded regions are the measured joints positions.}
  \label{fig:PhaseComp}
\end{figure}

The ability to successfully transfer the formal gaits generated to hardware is illustrated in \figref{fig:PhaseA} for Patient A, wherein the nominal (blue) and measured (shaded) trajectories are consistent throughout the experiment.
\add{The tracking error at the joint level for Patient A can be seen in \figref{fig:TrackingA}.} 
The motor torques resulting from tracking the nominal trajectories (cf. \figref{fig:Torques}) are also consistent with simulation. Note that motor-torque saturations are relatively uncommon as the gaits are designed to account for all hardware limits.
To compare the walking between patients, a representative selection of phase portraits for each patient are presented in \figref{fig:PhaseComp}; even though the gaits are not the same (as they have been generated to best suit each patient), they all display a common fundamental structure.  This is further illustrated in gait tiles of the patients walking in the exoskeleton (cf. \figref{fig:TileXPs}).

\subsection{Next steps}
While the dynamic walking gaits obtained are preliminary, and in no way constitute any kind of clinical evaluation, the ability to consistently realize them on patients points toward the validity of the framework being pursued. During the preliminary testing, the few tests that failed to completely traverse the lab were caused by foot scuffing or loss of lateral balance. In the next phase of testing, we will evaluate the G-HZD controllers discussed in the previous two sections.

%% file: section_conclusion.tex
\section{Future Work}
\label{sec:futurework}
While this paper presents an important accomplishment, a lot remains to be done in the field of actively controlled exoskeletons.  Future research directions involve developing control algorithms that \add{(a) directly address model uncertainty}; (b) support a rich set of behaviors such as standing and sitting and dynamic transitions between these; (c) enable push recovery and robustness to significant force disturbances that could arise due to contact with the environment or other humans; (d) use models of human comfort so as to balance between comfort and robustness of a gait of the exoskeleton; (e) capture human intent so as to enable human-driven autonomous exoskeleton control; (f) adapt to and provide user customized behaviors; and (g) improve the energy efficiency of assisted walking. \add{Regarding model uncertainty, it is hoped that methods developed in \cite{griffin2016nonholonomic} can be incorporated into the optimization and machine learning methods to provide robust nominal orbits.}

\begin{figure}
	\centering
    \includegraphics[width=1.0\columnwidth]{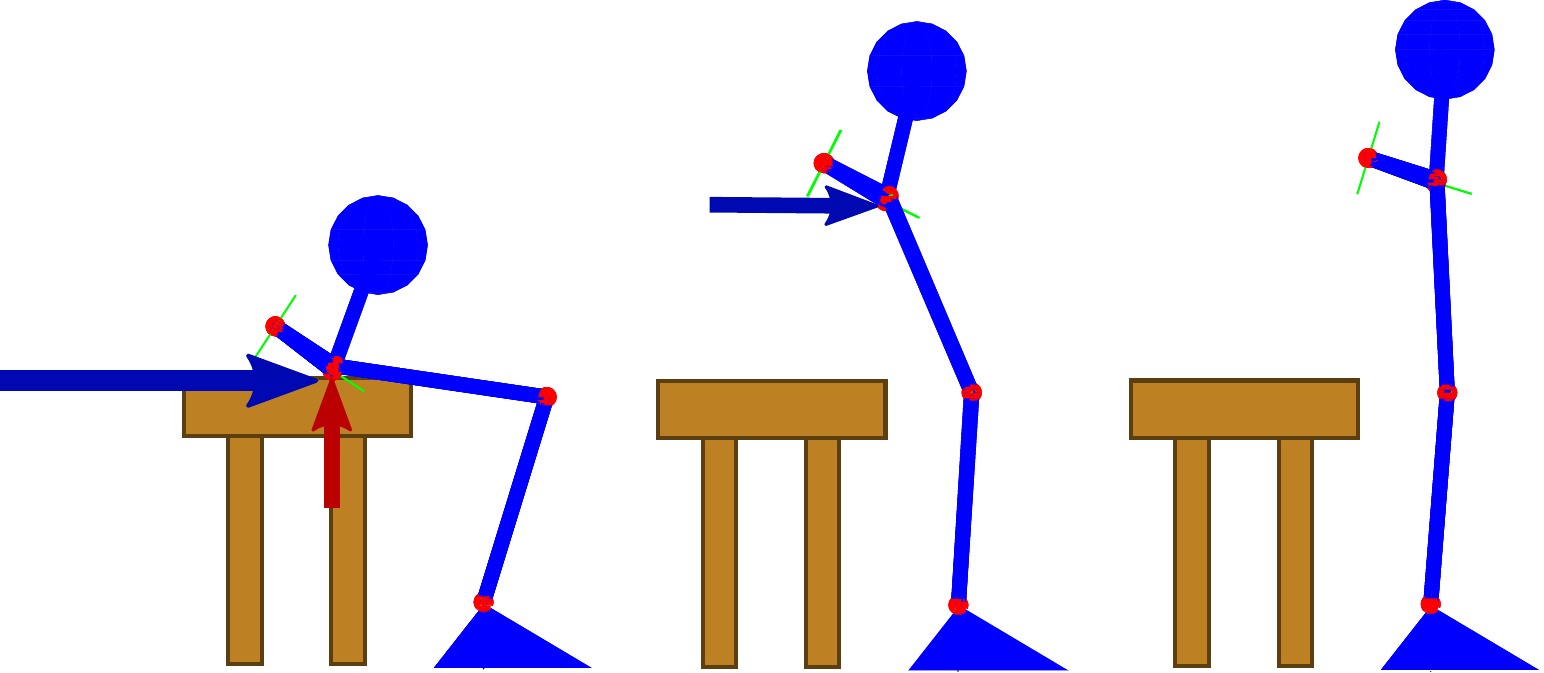}
	\caption{Stick figure animation of a user of the exoskeleton transitioning from sitting to standing. The first image is $t=0$, the second image is $t=0.6$~s, and the third image is $t=1.8$~s. The blue and red arrows show external forces applied to the system through the user's arms. In the optimization, it is assumed that the user can exert up to half their body weight as an external force.}
    \label{fig:standing_stick_anim}
\end{figure}

\begin{figure}[!t]
	\centering
    \includegraphics[width=1\columnwidth]{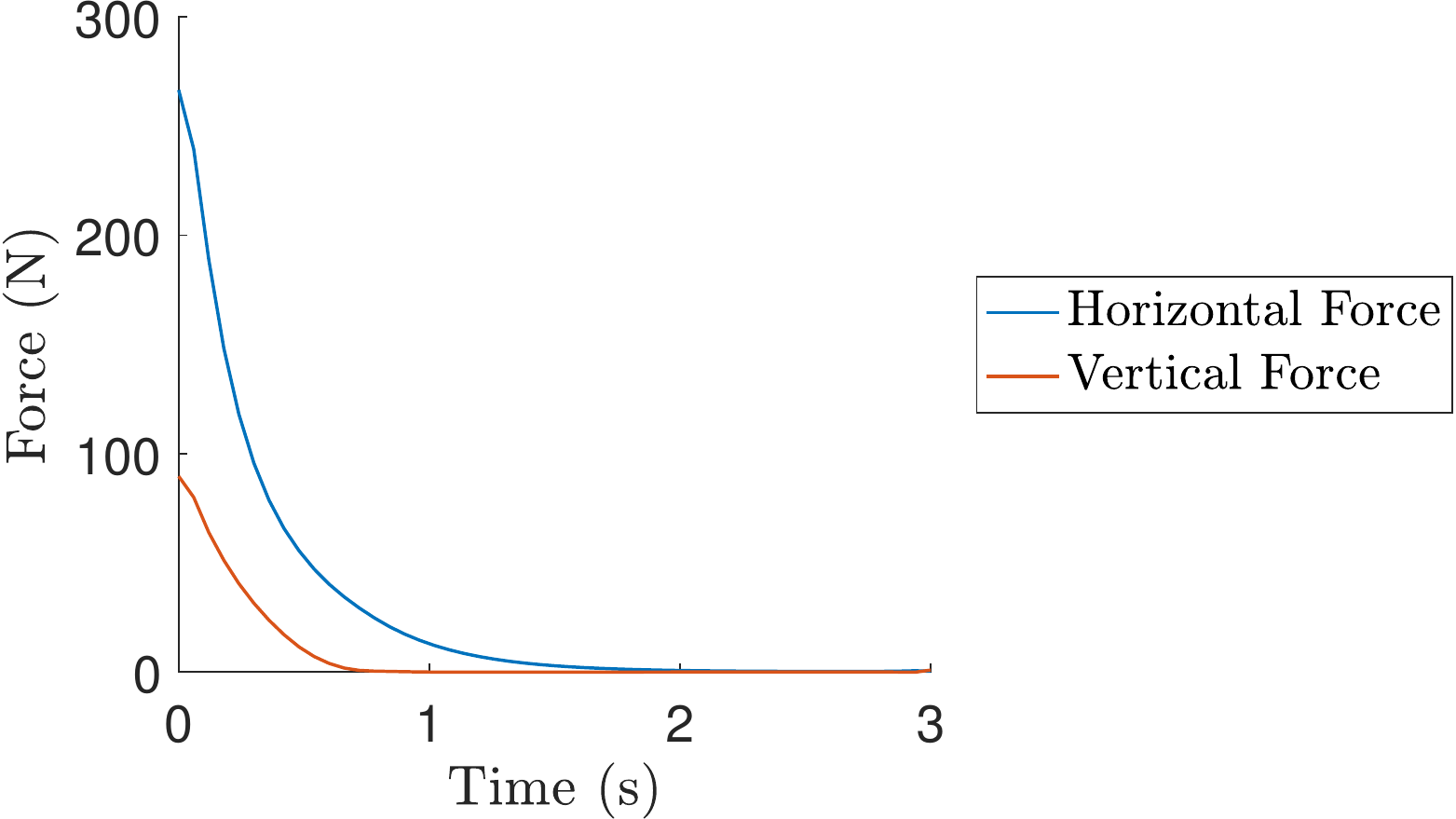}
	\caption{Force exerted by the user to assist in standing up. The exoskeleton is upright after 1.8 s from the start.}
    \label{fig:standing_external_force}
\end{figure}

We believe methods presented in this paper can be extended to tackle many of these tasks. For example, \figref{fig:standing_stick_anim} shows preliminary results for a transition from sitting to standing obtained via optimization. In the model, it was assumed that the user can use their arms to assist liftoff by applying an external force on the chair up to half their body weight. It can be seen from \figref{fig:standing_external_force} that most of the force is used to push the user horizontally off the chair.

\section{Summary Remarks}
\label{sec:conclusion}

The ATALANTE exoskeleton studied in this paper is the first to allow dynamic hands-free walking for paraplegics. Where other devices require crutches for lateral stabilization, the embedded control algorithms on the studied exoskeleton regulate leg motion so as to sustain a locally exponentially stable walking gait. The keys to realizing crutch-less dynamic walking were novel hardware, stiff enough to physically support a subject, powerful enough to move the device's legs quickly, and novel control mathematics developed over the past 15 years to allow bipedal robots to walk stably in uncertain environments and with imprecise dynamic models. 

The preliminary experimental results have demonstrated very slow walking on the order of 0.1 m/s. Stable gaits at 0.4 m/s have been achieved in simulation, and such speeds can be expected to be reached on the current hardware and with patients. New tools are coming online for the rapid computation of trajectories for high-degree-of-freedom mechanical systems as well as new control methods that are providing ways to mitigate the curse of dimensionality. A path forward to restoring locomotion for paraplegics is becoming more and more clear.

%% file: sidebar_pinned_vs_floating.tex
\subsection[Pinned versus Floating base models]{Pinned versus Floating base models}
\label{sidebar:pinned_vs_floating}

In Section \nameref{sec:model}, we use a set of floating base coordinates, attached to the pelvis of the exoskeleton system, to describe the configuration of the robot's base frame with respect to an inertial frame. The kinematic structure of the robot is then built relative to the pelvis base frame, branching into swing and stance legs.  

Another approach to describe the configuration of the robot is to use a \emph{pinned} open chain kinematic model \cite{Westervelt2007Feedback}.
In this approach, the robot's kinematic tree is built starting from the stance foot and branching into the torso and swing leg. It is assumed that the stance foot is attached to the ground through an ideal revolute joint. The advantage here is that the constraint forces enforcing the holonomic constraints on the stance foot no longer appear in the resulting equations of motion. 

While the two models are equivalent, note however, that the position of the swing foot with respect to the base frame (attached to the stance foot) in case of the pinned model involves more trigonometric (sine and cosine) terms as compared to the floating base model, where the base frame is attached to the pelvis. This is also reflected in the resulting equations of motion. One consequence is that optimizations for finding periodic orbits run faster for the floating-base model as compared to the pinned model.


%% file: sidebar_direct_collocation.tex
\subsection[How Direct Collocation Works]{How Direct Collocation Works}
\label{sidbar:direct_collocation}

\begin{figure}
	\begin{centering}
		\includegraphics[width=1\columnwidth]{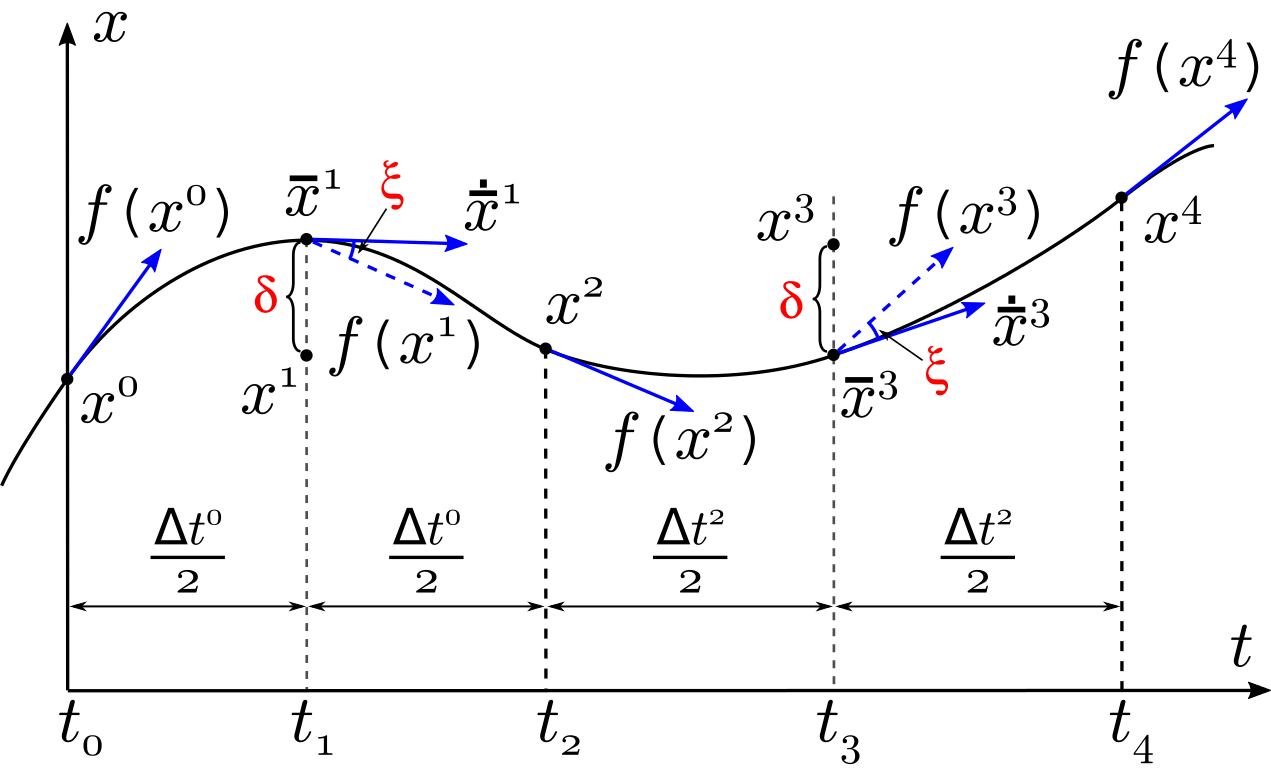}
	\end{centering}
	\caption{Illustration of defect constraints and node distribution of the direct collocation optimization \cite{Hereid2018Dynamic}.}
    \label{fig:direct_collocation}
\end{figure}

The direct collocation trajectory optimization utilizes the collocation methods for solving ordinary differential equations based on the finite step implicit Runge-Kutta methods \cite{Betts1998Survey}. Here, the specific formulation of the Hermite-Simpson algorithm, one of the most commonly used direct collocation schemes, is introduced. Consider a system of the form $\dot{x} = f(x,u)$, a trajectory optimization problem for such system can be stated as
\begin{align}
  J(x(t),u(t)) = \min_{u(t)}\quad & \int_0^T L(x(t),u(t)) dt \\
  \mathrm{st.}\quad &x(t) = \int_0^t f(x(t),u(t)) dt \nonumber \\
  & 0 \geq c(x(t), u(t)), \quad 0 \leq t \leq T, \nonumber
\end{align}
\add{where $L(\cdot)$ represent the running cost function, and $c(\cdot)$ represents the path constraints.} To solve this problem using direct collocation, a discrete representation of the continuous time solution is introduced. Specifically, the time interval $t \in [0,T]$ is divided into a fixed number of uniformly distributed intervals (see \figref{fig:direct_collocation}). In particular, the even-numbered nodes (e.g., $t_0$, $t_2$, $\dots$, $t_N$) are called cardinal nodes, and the odd-numbered nodes between every two cardinal nodes are called interior nodes. At each discrete node of $t = t_i$, an approximation of state variables $x_i = x(t_i)$ and control inputs $u_i = u(t_i)$ is introduced as a set of optimization variables to be solved. In this paper, the approximation of the slope of state variables $\dot{x}_i = \dot{x}(t_i)$ is also introduced as \emph{defect variables} in the optimization. 

The Hermite-Simpson methods then use piece-wise continuous cubic interpolation polynomials to approximate the solution of the system over each interval between two neighboring cardinal nodes. This approximation can be fully determined by the approximated state variables and slopes at the cardinal nodes. Hence, if the approximated states $x_i$ and slopes $\dot{x}_i$ at the interior nodes match the interpolation polynomial at time $t=t_i$ (e.g., $\bar{x}_i$ and $\dot{x}_i$ in \figref{fig:direct_collocation}), then the resulting piece-wise polynomials are considered as an approximated solution of the system \cite{Hargraves1987Direct}. To find this approximated solution, e.g., the discrete representation of the states, the original continuous time trajectory optimization problem can be converted to the following form given by
\begin{align}
  J(x_i, &u_i) = \min_{u_i}\quad \sum_{i=1}^{N-1} w_i L(x_i,u_i) \\
  \mathrm{st.}\quad &\dot{x}_i = f(x_i,u_i) \nonumber \\
  & c(x_i,u_i) \geq 0, \quad 0 \leq i \leq N \nonumber \\
  & \dot{x}_i - 3(x_{i+1} - x_{i-1})/2\Delta t_i + (\dot{x}_{i-1} + \dot{x}_{i+1})/4 = 0 \nonumber \\
  & x_i - (x_{i+1} + x_{i-1})/2 -  \Delta t_i (\dot{x}_{i-1} - \dot{x}_{i+1})/8 = 0,  \nonumber
\end{align}
where for all $i\in\{1,3,\cdots,N-1\}$, where $\Delta t_i = t_{i+1} - t_{i-1}$ is the time interval between two cardinal nodes, \add{and $w_i$ is weighting factor of each node determined by the Gaussian quadrature \cite{Hereid2018Dynamic}.} Specifically, the last two constraints are called \emph{collocation constraints}, which are determined by cubic interpolation polynomials. The above nonlinear programming problem can be solved straightforwardly by existing numerical NLP solvers.  

%% file: sidebar_cybalathon.tex
\subsection[The Cybathlon]{The Cybathlon}

\begin{figure*}[!t]
    \centering
    \begin{subfigure}[b]{0.82\textwidth}
        \centering
        \includegraphics[width=\textwidth, trim={0cm 25cm 0cm 0cm}, clip=true]{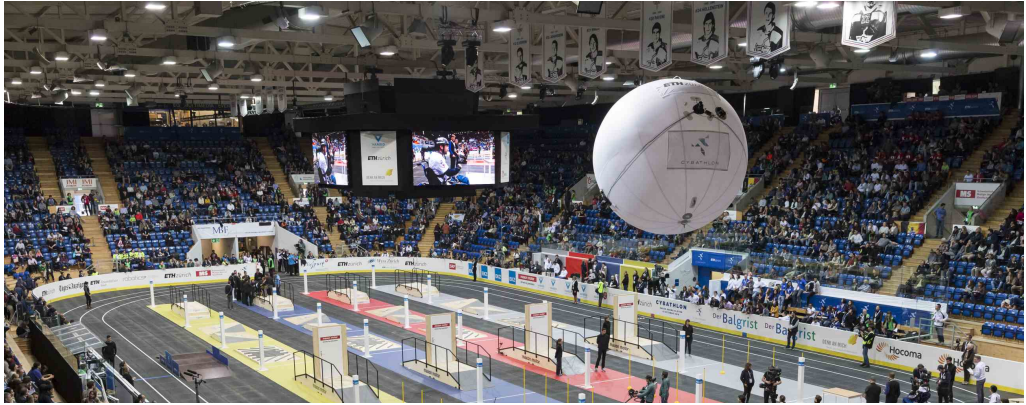}
        \caption{The exoskeleton race arena at Cybathlon 2016. (Source/image rights: ETH Zurich/Alessandro Della Bella)}
    \end{subfigure}%
    
    \begin{subfigure}[b]{0.4\textwidth}
        \includegraphics[width=\textwidth]{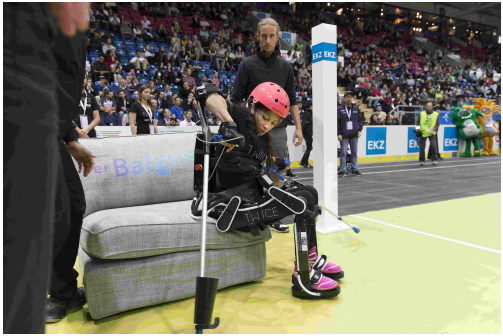}
        \caption{Sofa task. (Source/image rights: ETH Zurich / Nicola Pitaro)}
    \end{subfigure}
    ~
	\begin{subfigure}[b]{0.4\textwidth}
        \includegraphics[width=\textwidth]{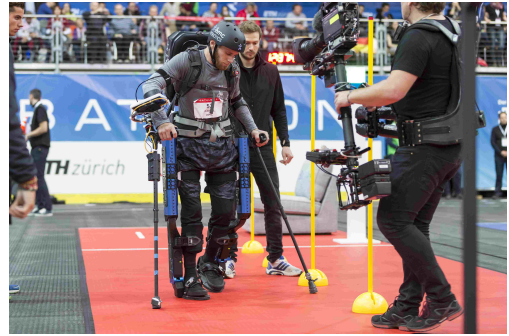}
        \caption{Slalom course. (Source/image rights: ETH Zurich / Alessandro Della Bella)}
    \end{subfigure}
         \begin{subfigure}[b]{0.4\textwidth}
        \includegraphics[width=\textwidth]{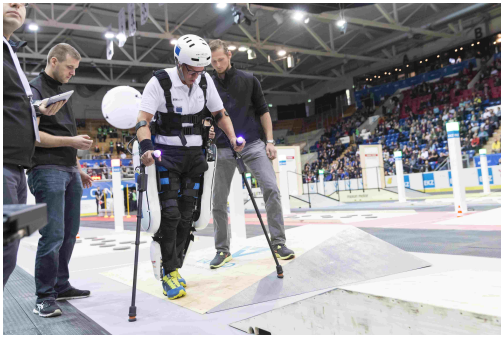}
        \caption{Walking across tilted paths. (Source/image rights: ETH Zurich / Alessandro Della Bella)}
    \end{subfigure}
            ~
\begin{subfigure}[b]{0.4\textwidth}
        \includegraphics[width=\textwidth]{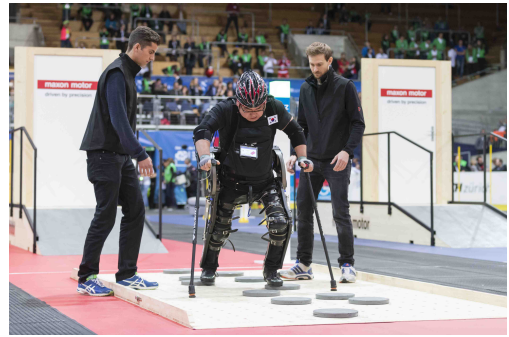}
        \caption{Stepping stones task. (Source/image rights: ETH Zurich / Alessandro Della Bella)}
    \end{subfigure}
    
    \caption{Exoskeleton Race at Cybathlon 2016. Images obtained from \cite{CybathalonImages}.}
\end{figure*}

In October of 2016, ETH Zurich hosted a one-of-a-kind race for people with disabilities using advanced assistive devices. The goal of the event was to enhance public awareness about the challenges faced by people with disabilities. The event also served as a common platform for technical exchange between various research organizations and companies that develop assistive biomechatronic devices \add{for} people with disabilities.  A total of 66 pilots, 56 teams, 25 nations, and 400 team members participated in various races across six different disciplines including powered arm and leg prosthesis, brain-computer interface, powered wheelchairs and exoskeletons \cite{riener2014cybathlon}. 

The powered exoskeleton race was intended for participants with complete thoracic or lumbar spinal cord injuries (SCI). Pilots, equipped with exoskeletons, were asked to complete as many tasks as possible in a ten-minute time frame. These tasks were representative of common day-to-day activities such as sitting on and standing up from a chair, walking around obstacles (slalom course), walking over ramps and navigating through doorways, walking across tilted paths and over discrete footholds (stepping stones). The race saw participation from nine different teams. The German team, ReWalk, won first place, followed by IHMC from the United States, and SG Mechatronics from Republic of Korea. The next Cybathlon is scheduled for May 2020.